\pdfoutput=1

\documentclass[11pt]{article}

\usepackage[final]{acl}

\usepackage{times}
\usepackage{latexsym}

\usepackage[T1]{fontenc}

\usepackage[utf8]{inputenc}

\usepackage{microtype}

\usepackage{inconsolata}

\usepackage{amsfonts} 
\usepackage{amssymb}  
\usepackage{pifont}
\usepackage{arydshln} 
\usepackage{xcolor} 
\usepackage{colortbl}
\usepackage{mathtools} 
\usepackage{adjustbox} 
\usepackage{multirow} 
\usepackage{paralist} 
\usepackage{CJKutf8} 
\usepackage{booktabs} 

\usepackage{enumerate}
\usepackage{bbding} 
\usepackage{wasysym}
\usepackage{caption}
\usepackage{subcaption}
\usepackage{cleveref}
\usepackage[ruled,linesnumbered]{algorithm2e}
\SetKwInOut{Parameter}{Parameters}

\usepackage{bm}
\usepackage{scalefnt}
\usepackage{pifont}

\newcommand{\method}{\texttt{UltraTool}\xspace}
\title{Planning, Creation, Usage: Benchmarking LLMs for Comprehensive \\ Tool Utilization in Real-World Complex Scenarios}

\newcommand{\samethanks}[1][\value{footnote}]{\footnotemark[#1]}
\author{
	Shijue Huang$^{1,5}$\thanks{~~This work was done during the internship at Huawei
		Noah’s Ark Lab.}
	Wanjun Zhong$^{3}$\thanks{\,\, Corresponding authors.}
	Jianqiao Lu$^{4}$
	Qi Zhu$^{3}$
	Jiahui Gao$^{3}$
	Weiwen Liu$^{3}$ \\
	\textbf{Yutai Hou}$^{3}$
	\textbf{Xingshan Zeng}$^{3}$
	\textbf{Yasheng Wang}$^{3}$
	\textbf{Lifeng Shang}$^{3}$
	\textbf{Xin Jiang}$^{3}$ \\
	\textbf{Ruifeng Xu}$^{1,2,5}$\samethanks
	\textbf{Qun Liu}$^{3}$\\
	$^{1}$Harbin Institute of Technology, Shenzhen, China
	$^{2}$Peng Cheng Laboratory, Shenzhen, China\\
        $^{3}$Huawei Technologies Co., Ltd
	$^{4}$The University of Hong Kong \\
 $^{5}$Guangdong Provincial Key Laboratory of Novel Security Intelligence Technologies\\
 {\tt joehsj310@gmail.com,}
 {\tt zhongwanjun1@huawei.com,}
 {\tt  xuruifeng@hit.edu.cn}\\
}

\begin{document}
\maketitle

\begin{abstract}
The recent trend of using Large Language Models (LLMs) as tool
agents in real-world applications underscores the necessity for comprehensive evaluations of their capabilities, particularly in complex scenarios involving planning, creating, and using tools.
However, existing benchmarks typically focus on simple synthesized queries that do not reflect real-world complexity, thereby offering limited perspectives in evaluating tool utilization.
To address this issue, we present \method, a novel benchmark designed to improve and evaluate LLMs' ability in tool utilization within real-world scenarios. \method focuses on the entire process of using tools - from planning and creating to applying them in complex tasks. 
It emphasizes real-world complexities, demanding accurate,
multi-step planning for effective problem-solving. 
A key feature of \method is its independent evaluation of planning with natural language, which happens before tool usage and simplifies the
task solving by mapping out the intermediate steps.
Thus, unlike previous work, it eliminates the restriction of pre-defined toolset. 
Through extensive experiments on various LLMs, we offer novel insights into the evaluation of capabilities of LLMs in tool utilization, thereby contributing a fresh perspective to this rapidly evolving field.
The benchmark 
is publicly available at
\url{https://github.com/JoeYing1019/UltraTool}.

\end{abstract}

\section{Introduction}
\label{Introduction}
\begin{figure}[t]
	\centering
		\centering
		\includegraphics[width=0.45\textwidth]{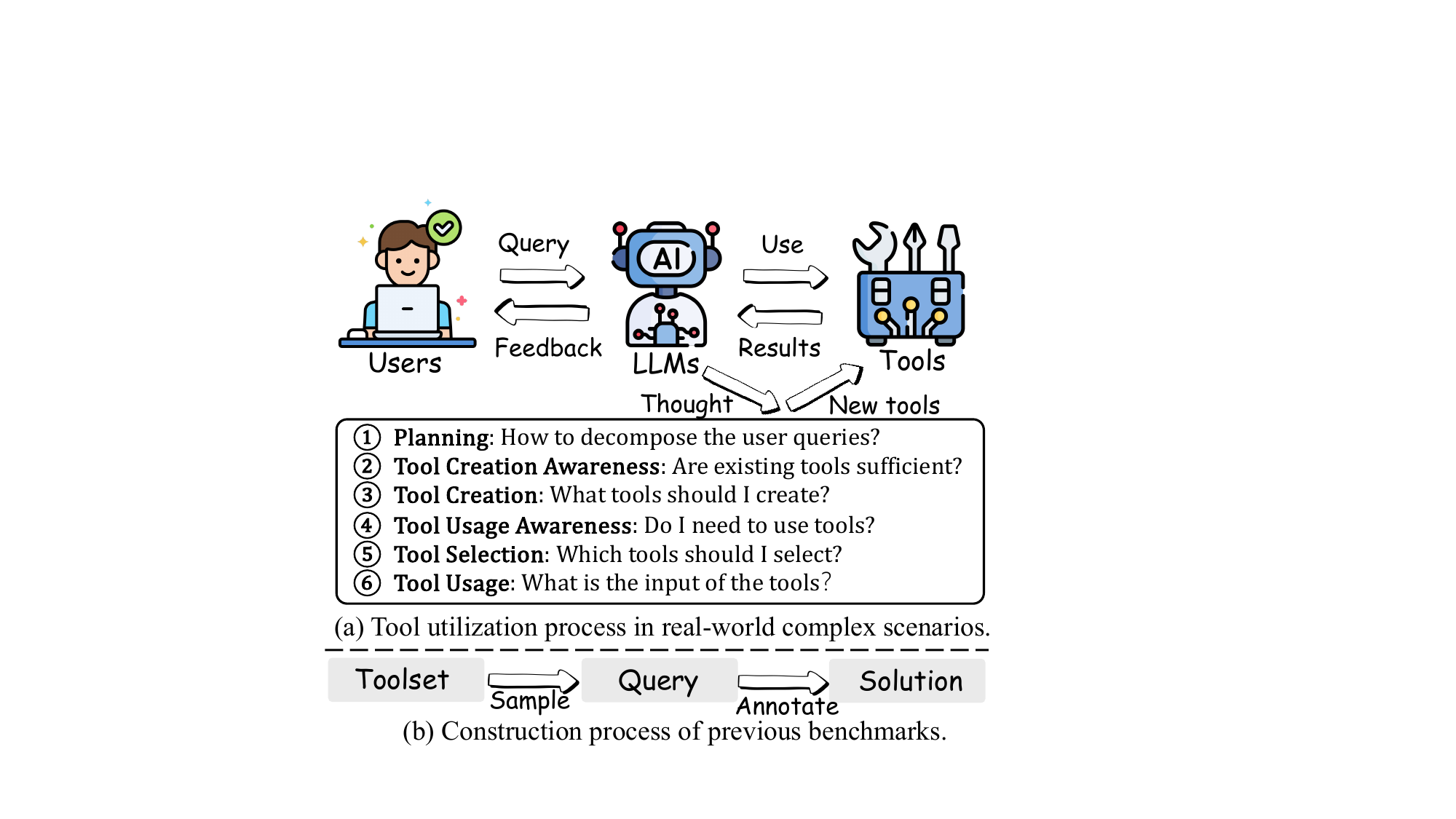}
		\caption{Illustration of (a) tool utilization process in real-world complex scenarios and (b) construction process of previous benchmarks.
		 }
	\label{fig:intro}
 \vspace{-0.2cm}
\end{figure}

\begin{table*}[t]
\centering
\begin{adjustbox}{width=0.98\textwidth}
\begin{tabular}{lcccccccc}
\toprule
\multirow{2}{*}{Resource} 
& ToolAlpaca & APIBench & APIBank & ToolBench &  MetaTool & UltraTool \\ 
&\citet{tang2023toolalpaca} & \citet{patil2023gorilla} & \citet{li-etal-2023-api}&\citet{qin2023toolllm}  & \citet{huang2023metatool}& (Ours) & \\ 
\midrule

Evaluation Range 
& \ding{177}& \ding{177} & \ding{177} & \ding{177} & \ding{175}\ding{176} & \ding{172}\ding{173}\ding{174}\ding{175}\ding{176}\ding{177} \\

Real-world Query 
& \ding{55} & \ding{55}& \ding{55} & \ding{55}& \ding{55}& \ding{51}\\

Multi-Tool Test 
& \ding{55}& \ding{55} & \ding{55} & \ding{51}& \ding{51}& \ding{51}\\

Different Domains 
& \ding{55} & \ding{55} & \ding{51}& \ding{51}& \ding{51} & \ding{51}\\
\bottomrule
\end{tabular}
\end{adjustbox}
\caption{Comparison of previous benchmarks and \method.
}
\vspace{-0.2cm}
\label{tab:benchmark_comparison}
\end{table*}

Recent advancements in equipping Large Language Models (LLMs) ~\citep{du2022glm, touvron2023llama, vicuna2023, bai2023qwen} with external tools ~\citep{patil2023gorilla, schick2023toolformer, qin2023tool} have markedly improved capability of AI systems in solving complex real-world tasks. 
As this field evolves, it becomes crucial to conduct a comprehensive evaluation covering full aspects of tool utilization, particularly within complex real-world contexts. As depicted in Figure~\ref{fig:intro} (a), addressing real-world tasks often necessitates not only the planning and usage of multiple tools but also the creation of new tools, when the existing tools are not enough to meet all specific requirements.
However, existing benchmarks
~\citep{tang2023toolalpaca,xu2023tool, patil2023gorilla,li-etal-2023-api,qin2023toolllm,huang2023metatool}
often focus only on limited dimensions 
of this entire process. Besides, the user queries in most existing benchmarks exhibit limitations in mirroring the complexity of real-world tasks and unreasonable dependency on pre-defined toolsets. 

To tackle these challenges, we introduce \method, a benchmark covering a wide range of evaluation dimensions. 
It is constructed on complex, real-world queries and involves evaluation of tool-independent natural-language planning and advanced tool creation capabilities. 
\method comprises 5,824 examples, spanning 22 diverse domains and incorporating 2,032 tools, and it
comprehensively evaluates the tool utilization process including six
dimensions covering three aspects:
 \ding{172} \textbf{Planning}: decomposing the complex goal into logical sequence of simpler sub-tasks for effective problem-solving.
 \textbf{Tool Creation} comprises two dimensions:
 \ding{173} awareness - assessing whether existing tools suffice, and \ding{174} creation - developing the necessary tools if existing ones are inadequate. \textbf{Tool Usage} involves three dimensions:
 \ding{175} awareness - determining which sub-tasks require tools, \ding{176} selection - choosing the appropriate tools, and \ding{177} usage - specifying the input parameters for these tools.
As demonstrated in \tablename~\ref{tab:benchmark_comparison}, recent benchmarks in tool utilization tend to focus only on limited dimensions.
For examples, ~\citet{tang2023toolalpaca, patil2023gorilla,li-etal-2023-api, qin2023toolllm} focus on tool usage (\ding{177}), while
\citet{huang2023metatool} 
evaluates tool usage awareness (\ding{175}) and tool selection (\ding{176}).
Comparing with them, we further evaluate the capability in planning (\ding{172}) and tool creation (\ding{173} and \ding{174}), which is crucial for LLMs to adeptly navigate complex and varied real-world user demands.

Moreover, the queries in \method have better realism and complexity.
The natural workflow of tool-augmented task solving includes: ``user queries LLMs $\rightarrow$ LLMs pose a solution plan $\rightarrow$ LLMs create/select tools to solve sub-task within the plan''. 
Importantly, the query and the plan should not be constrained by pre-existing tools.
In contrast, existing benchmarks~\citep{qin2023toolllm, li-etal-2023-api, huang2023metatool} are typically constructed by 
collecting toolset, generating simulated queries with tools randomly selected from toolset, and then
annotating the solution as shown in \figurename~\ref{fig:intro} (b).
Despite the collecting efficiency, the generated queries may not accurately mirror real user demands. 
Furthermore, the randomly selected tools may lack a coherent logical relationship, potentially leading to skewed influences on the subsequent process.
To ensure the complexity and reality of queries, we collect real-world multi-domain user queries with high complexity. 
 Specifically, we engage experts from various domains to craft complex queries that reflect real-world needs and potentially incorporate the use of diverse tools.

A distinct feature of \method is it explicitly evaluates natural language (NL)-based plan, which simplifies task solving by decomposing a complex goal into several simpler sub-tasks described in NL. 
Each sub-task is then solved by tool creation and usage. 
Therefore, prior planning overcomes the restriction brought by limited pre-existing tools.

We conduct experiments on various LLMs and make in-depth analyses 
about strengths and challenges in the tool utilization of LLMs.
Contributions of this work are:
(1) \method is a comprehensive evaluation benchmark derived from complex real-world queries, covering six key dimensions
in tool utilization.
(2) \method explicitly evaluates NL-based planning and advanced tool creation capabilities. 
(3) Extensive experiments uncover the limitations and inspire the future direction of LLMs in tool utilization.

\section{\method Construction}

\label{sec:dataset-construction}

\begin{figure*}[t]
	\centering
		\centering
		\includegraphics[width=\textwidth]{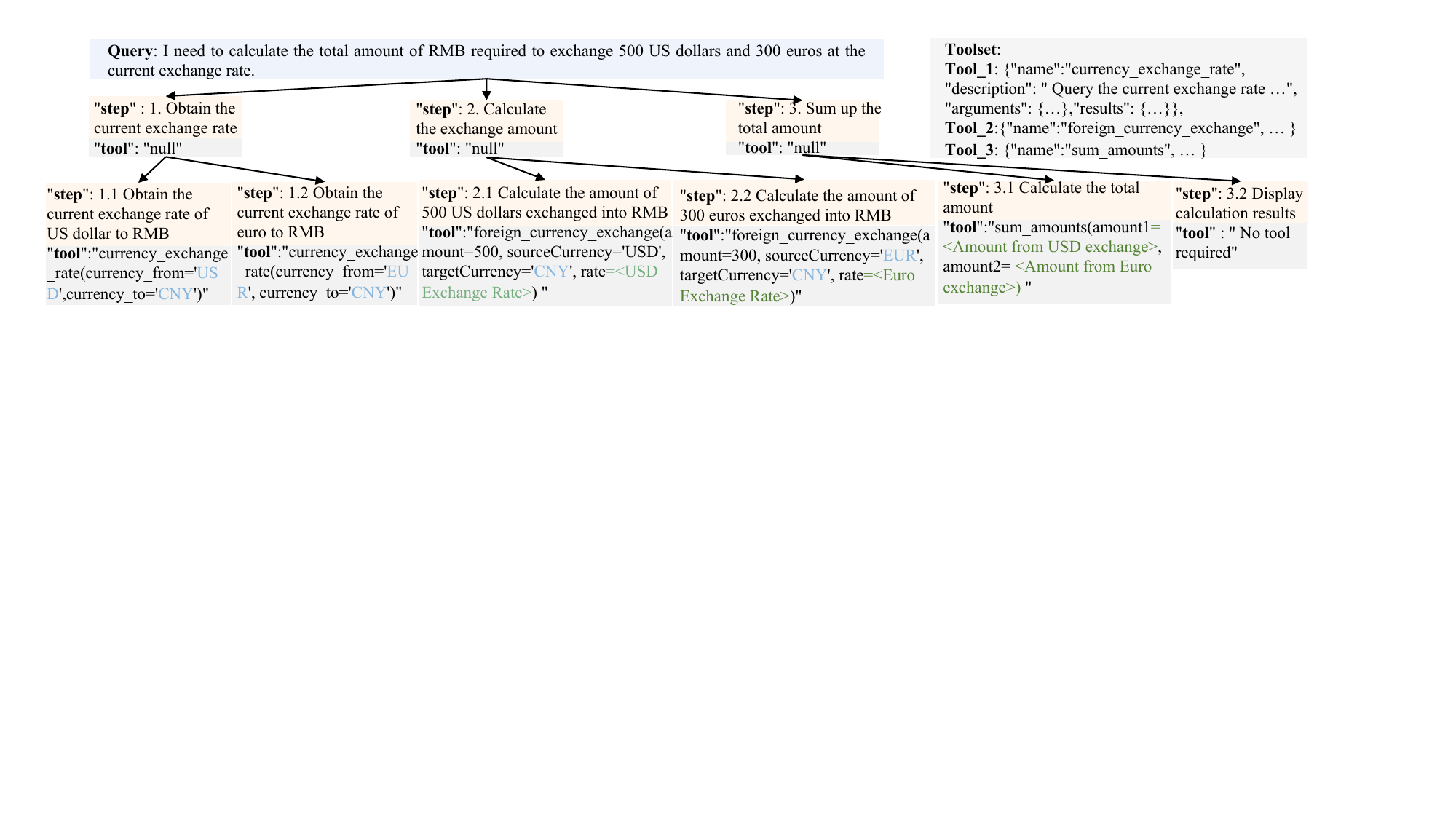}
		\caption{Data example of \method, including user query, tree-structure planning, and toolset.
		 }
	\label{fig:example}
 \vspace{-0.2cm}
\end{figure*}

\method begins by gathering real-world user queries and encompasses a comprehensive range of evaluation dimensions, including planning, tool creation and tool usage. 
As depicted in \figurename~\ref{fig:overall}, the construction process of \method includes:
(1) query collection ($\S \ref{sec:query collection}$); (2) solution annotation ($\S \ref{sec:plan}$); and (3) manual refinement ($\S \ref{sec:human}$).
For the detailed prompts utilized in the \method construction process, please refer to Appendix \ref{sec: data prompt}.

\subsection{Definition}
To formalize, a triple $(Q, P, T)$  is regarded as a sample within \method.
Specifically, $Q$ represents the user query, while $P=[(s_1, t_1), (s_2, t_2), ... ,(s_n, t_n)]$ is the NL-based plan, 
In this context, each element $(s_i, t_i)$ comprises a step $s_i$ and an associated tool calling message $t_i$.
Furthermore, $T=[tool_1, tool_2, ... ,tool_m]$ is the corresponding toolset. 
As depicted in \figurename~\ref{fig:example}, 
$P$ maintains a hierarchical tree structure
, which includes both ancestral steps (e.g., 1., 2., ...) and child steps (e.g., 1.1, 1.2, ...).
The child steps fall into two distinct categories: 
 (1) the \textit{tool-free step}, which the LLMs can infer without requiring tool callings, distinguished by the tool calling message $t_i$ as "No tool required"; and (2) the \textit{tool-usage step}, which necessitates calling tools to complete the task, indicated by the tool callings message
 $t_i$ specifying the tool name $tool_i^{\text{name}}$ and the necessary arguments 
 $[\{args_j^{\text{name}}:args_j^{\text{value}}\}]_{j=1}^{i_k}$.
Here, $i_k$ denotes the number of arguments, $args_j^{\text{name}}$ is the name of an argument belonging to the defined argument properties of $tool_i$, and $args_j^{\text{value}}$ is the argument value derived from the query context $Q$ or the outputs of preceding tool callings.

\subsection{Query Collection}\label{sec:query collection}
\begin{figure}[t]
	\centering
	\includegraphics[width=0.48\textwidth]{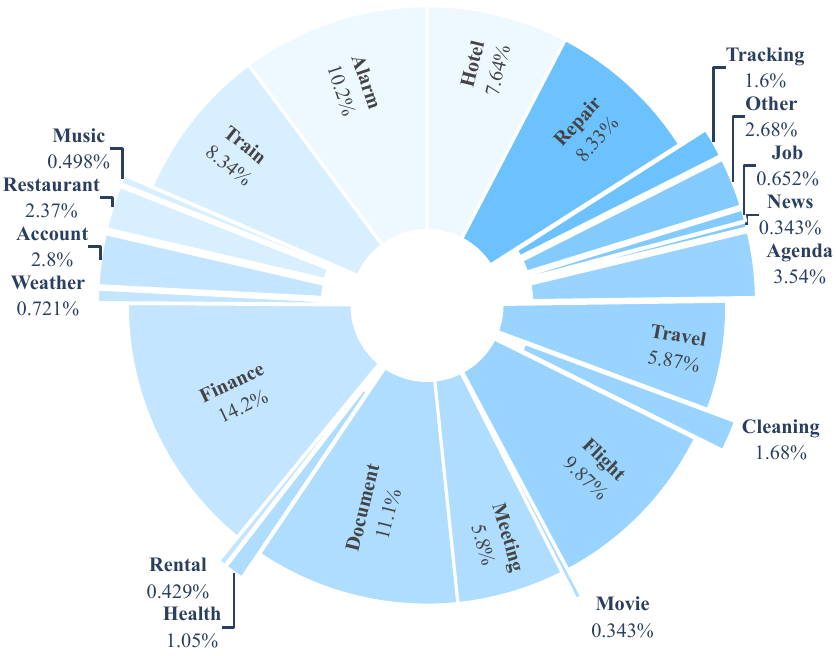}
		\caption{Specific domain distribution of \method.
		 }
	\label{fig:stat}
   \vspace{-0.2cm}
\end{figure}

To ensure comprehensive coverage across various domains and meet diverse requirements in real-world scenarios, 
we carefully selected more than 20 domains such as \textit{Alarm, Train, Flight, Hotel}, etc, as shown in \figurename~\ref{fig:stat}.
In detail, we collaborate with a wide array of domain-specific experts to gather complex, de-identified user queries that necessitate complex tool usage as depicted in \figurename~\ref{fig:overall} (a). 
To better align each query with its corresponding tool usage, we also request that the experts suggest several potential tools for each query. These suggestions are formatted in accordance with the style of ChatGPT plugins~\footnote{\url{https://platform.openai.com/docs/guides/function-calling}}, which aided in crafting an initial toolset for each query.
The query collection guideline can be found in Appendix~\ref{sec:query_guideline}.

As shown in \figurename~\ref{fig:overall} (a), to create more diverse and challenging queries, we utilized GPT-4 to generalize and complicate the initial queries. 
The examples of the generalization and complication processes are listed in Appendix~\ref{sec:examples}.
We then merge the original, generalized, and complicated queries into a comprehensive collection. 
Additionally, every query undergoes a manual review to guarantee quality and applicability, and queries that do not align with actual human needs are removed.

\subsection{Solution Annotation}\label{sec:plan}

Due to the complexity of the queries addressed by \method, 
\method employs a multi-step plan as the backbone of solution for effective problem-solving, wherein each step includes a corresponding tool calling message. By breaking down user queries into simpler tasks via this planning process, these tasks can subsequently be executed by the designated tools. This strategy is not only in line with the problem-solving logic of humans but also reduces the complexity of tool usage, thereby facilitating the handling of more challenging tasks.
We introduce an automatic solution annotation approach utilizing GPT-4, including: (1) plan annotation; (2) tool creation and plan refinement; (3) tool calling message annotation; and (4) tool merge.

 \begin{figure*}[t]
	\centering
	\includegraphics[width=0.93\textwidth]{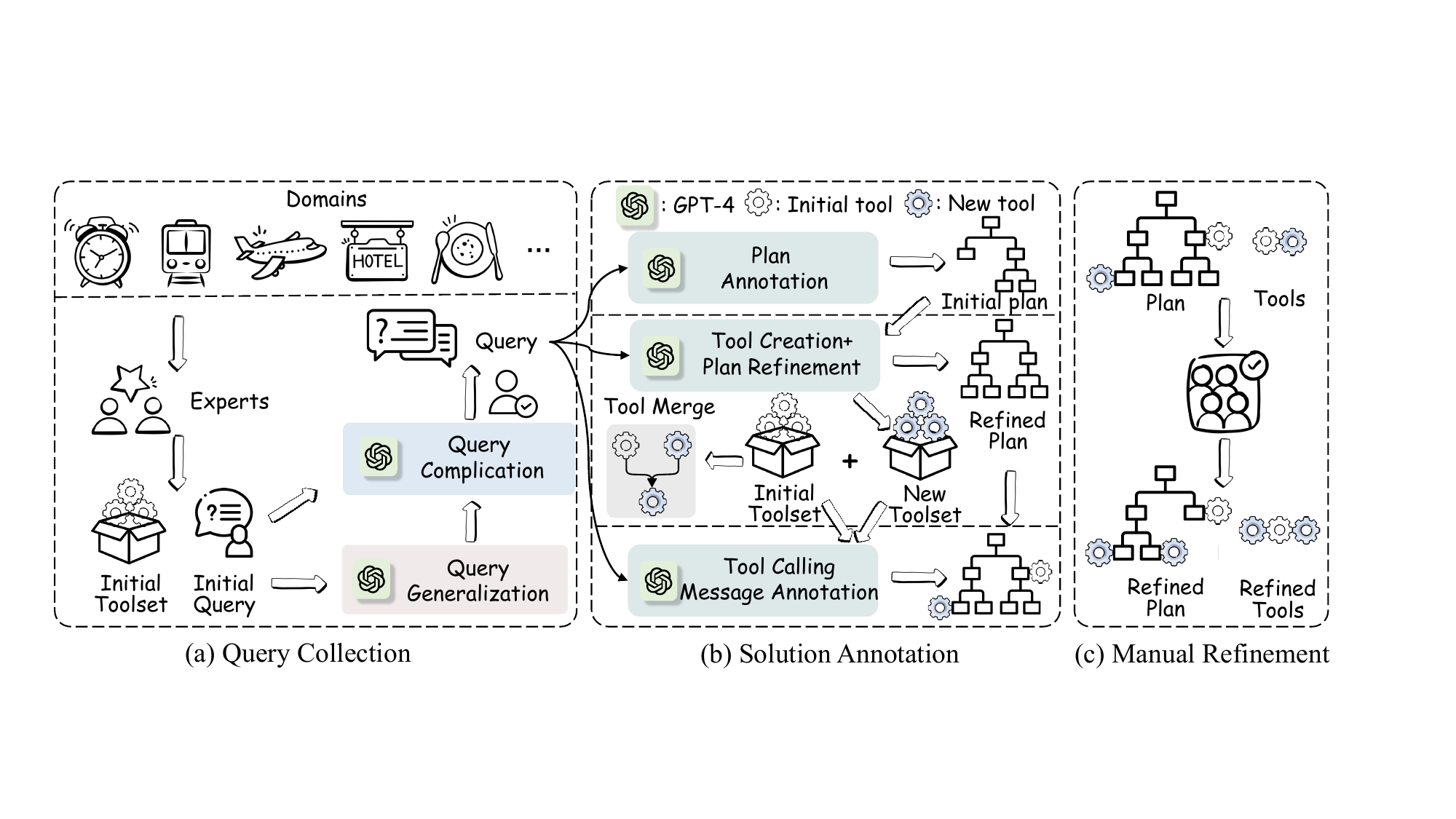}
	\caption{
 The overall construction process of \method,
 including (a) query collection, (b) solution annotation, and (c) manual refinement.
	}
	\label{fig:overall}
  \vspace{-0.2cm}
\end{figure*}

\noindent \textbf{Plan Annotation.}
As illustrated in \figurename~\ref{fig:overall} (b), 
upon receiving a user query, GPT-4 is utilized to formulate an initial multi-step, tree-structured plan $[s_1, ..., s_n]$.
This fundamental plan is deliberately crafted without considering a predefined toolset. Such a strategy ensures a primary focus on task decomposition and the structural integrity of the plan.
Moreover, it adeptly avoids the potential limitations that could surface when the initial toolset does not fully meet the requirements of user query.

\noindent \textbf{Tool Creation and Plan Refinement.}
The initial toolset may not meet all requirements within the complex user queries in \method, and the processes of generalization and complication may lead to emergence of new tool usage demands. Therefore, as depicted in \figurename~\ref{fig:overall} (b), to accommodate the need for tools not present in initial toolset, we utilize GPT-4 to discern whether need to create new tools
and create
new tools in prescribed format.

Subsequently, we implement an automatic refinement process to enhance the quality of the NL-based plan. Given that the existing toolset has already fulfilled all the requirements of the user query, we provide GPT-4 with the query, the initial plan, and the comprehensive toolset (which includes both the initial and the newly created tools). This allows GPT-4 to refine the plan further in terms of the comprehensiveness of information included and the compatibility with tool callings.

\noindent \textbf{Tool Calling Message Annotation.}
As shown in \figurename~\ref{fig:overall} (b), 
we utilize GPT-4 to annotate the tool calling messages $[t_1, ..., t_n]$.
In detail, given the query, the refined plan, and the comprehensive toolset, GPT-4 initially differentiates between the \textit{tool-free steps} and \textit{tool-usage steps}.
Subsequently, for each \textit{tool-usage step}, GPT-4 selects the most appropriate tool from the comprehensive toolset. Moreover, utilizing the user's query and the contextual information within the plan, GPT-4 generates the necessary arguments for the chosen tool.

It is also noteworthy that \method incorporates nested tool callings, wherein the output from one tool serves as a requisite parameter for another subsequent tool. This feature significantly enhances our benchmark's alignment with the complexities encountered in real-world scenarios.

\noindent \textbf{Tool Merge.}
Due to each tool being created separately for the queries, there may be similar tools that exist in the whole benchmark. To address this issue, we manually merge groups of tools that have similar functions into a single tool. Specifically, we combine the arguments and results fields of similar tools and remove any duplicate properties, then we write an appropriate name and description for the merged tool. The remained toolset of \method contains 2,032 distinct tools.

\subsection{Manual Refinement} \label{sec:human}
To guarantee and further enhance the data quality within \method, we conduct a careful manual refinement process for all samples, as illustrated in \figurename~\ref{fig:overall} (c). This process includes eliminating redundant steps, supplementing missing steps for coherent planning logic, rectifying tool usage demands and inappropriate tool selections, completing and refining the tool calling messages, and deleting data of substandard quality. We employ six experts endowed with specialized knowledge to refine the data,
and we ensure that all the polished data undergoes a double-check process to maintain high-quality standards. 
The refinement guideline is provided in Appendix~\ref{sec:refine_guideline}.

\subsection{Data Summary}
Ultimately, we construct the \method, which encompasses 22 domains, 2,032 tools, and 5,824 samples, detailed statistics are presented in Appendix~\ref{sec:data_stat}. And the source language of our collected data is Chinese, to broaden the scope of evaluation, we translate it to English version through GPT-4 and manual refinement. Therefore, \method supports two languages: Chinese and English, represented by the Chinese-dataset and the English-dataset.

\section{\method Evaluation}
In this section, we introduce the metrics ($\S \ref{sec:metrics}$), and the definition and measurement of each evaluation aspect, including: planning($\S \ref{sec:planning}$), tool creation($\S \ref{sec:tool_creation}$) and tool usage($\S \ref{sec:tool_usage}$).

\subsection{Metrics} \label{sec:metrics}
Three evaluation metrics are employed within \method, comprising:

\noindent (i) \textit{Multi-Dimensional Point-Wise LLM-as-Judge Method}~\citep{liu2023alignbench}, which
utilizes a LLM-scorer as automatic evaluator. By defining task-specific evaluation dimensions $M=[d_1, d_2, ...d_m]$ and providing the query $Q$, the model’s response $p$, and the reference answer $y$,
the LLM-scorer is prompted to provide a multi-dimensional score, along with an overall score:
\begin{eqnarray}
\bm S = \text{LLM-scorer}(M, [Q, p, y]), 
\end{eqnarray}
where $\bm S =[s_{d_1}, ... , s_{d_m},s_{\text{overall}}]$ contains dimension scores $s_{d_1}, ... , s_{d_m}$ and the final overall score $s_{\text{overall}}$, and all the scores are ranging from 1 to 10.
The detailed prompts are listed in Appendix~\ref{sec:eval prompt}.

\noindent(ii) \textit{Key-Value based Accuracy} and (iii) \textit{Key-Value based Levenshtein Distance}, which are variants of traditional accuracy and Levenshtein distance~\citep{10.5555/1822502} metrics. Given the key-value format model’s response $(p_k, p_v)$ and reference answer $(y_k, y_v)$, where the keys $p_k$ and $y_k$ represent steps and values $p_v$ and $y_v$ denote task-specific results,
these metrics compute the accuracy or normalized Levenshtein distance between the values
when the keys 
match:
\begin{eqnarray}
\bm S =
 \begin{cases}
 F(p_v, y_v) & \text{ if } p_k = y_k \\
 0 & \text{ if }  p_k \neq y_k
\end{cases},
\end{eqnarray}
where $\bm S$ is the computated score and $F$ represents the calculation function, which can applied by
accuracy or normalized Levenshtein distance.

The multiple-step nature of the plan resulting in multiple predictions for a sample, 
so these metrics may be computed at global level that evaluates multiple predictions together or local level that evaluates each prediction separately. 
More illustration of computation level can be found in Appendix~\ref{sec:level}.

\subsection{Planning} \label{sec:planning}
Given a query $Q$, LLMs need to decompose the user query and generate a step sequence in hierarchical form $[p_1^{\text{PL}}, p_2^{\text{PL}}, ... ,p_n^{\text{PL}}]$
, to serve as the plan. 
Here, each $p_i^{\text{PL}}$ represents a step. 
We utilize the \textit{Multi-Dimensional Point-Wise LLM-as-Judge Method}, applied at a global level, to evaluate LLMs' planning ability.
The generated plan is evaluated across six score dimensions including \textit{Accuracy}, \textit{Completeness}, \textit{Executability}, \textit{Syntactic Soundness}, \textit{Structural Rationality} and \textit{Efficiency}, detailed dimension definitions are listed in Appendix~\ref{sec: plan_dimensions}.

\subsection{Tool creation}\label{sec:tool_creation}
Since many real-world demands cannot be addressed by existing real-world tools, we design our tools as tool skeletons that contain all the necessary information for calling, as illustrated in \figurename~\ref{fig:example}. Despite these skeletons are not specific implementations, they serve as simulated tools that accurately represent the tool’s functionality and provide guidance for the future development of actual tools.

Furthermore, given the golden plan $P$ and corresponding toolset $T$, to evaluate whether the LLM can accurately be aware that the provided toolset is sufficient and effectively create the tools that are lacking, we construct an alternative toolset $\hat{T}$ that may not contain all the necessary tools.

\noindent \textbf{Awareness.}
Given plan $P$ and toolset $\hat{T}$,
LLMs need to determine whether each \textit{tool-usage step} in the plan can be matched with an appropriate tool from $\hat{T}$.
The output is a predicted sequence $[(s_i,p_i^{\text{TCA}}), ... ,(s_j,p_j^{\text{TCA}})]$, where $p_i^{\text{TCA}} \in \{0, 1\}$ indicates the matching availability of a suitable tool within $\hat{T}$ for step $s_i$. 
The evaluator of tool creation awareness is conducted through the \textit{Key-Value based Accuracy} metric, which is calculated at both the global and local levels.

\noindent \textbf{Creation.}
Given plan $P$ and toolset $\hat{T}$, LLMs are required to create the lacking tool in required format for those \textit{tool-usage steps} that can not match a suitable tool in $\hat{T}$. 
The output is a predicted sequence $[(s_i,p_i^{\text{TC}}), ... ,(s_j,p_j^{\text{TC}})]$, where $p_i^{\text{TC}}$ denotes the created tool for step $s_i$.
The \textit{Multi-Dimensional Point-Wise LLM-as-Judge Method}, calculated at the global level, is adopted as the evaluator for tool creation. We evaluate the newly created tool across five score dimensions including \textit{Format Compliance}, \textit{Accuracy}, \textit{Content Reasonableness}, \textit{Executability} and \textit{Richness}, corresponding dimension definitions can be found in Appendix~\ref{sec: tool_dimensions}.

\subsection{Tool Usage}\label{sec:tool_usage}
\noindent \textbf{Awareness.}
Given the plan $P$, LLMs need to determine whether child steps require the usage of tools and output a predicted sequence $[(s_i,p_i^{\text{TUA}}), ... ,(s_j,p_j^{\text{TUA}})]$, where $p_i^{\text{TUA}} \in \{0, 1\}$ indicates whether step $s_i$ need to use tool.
The tool usage awareness evaluator is \textit{Key-Value based Accuracy} calculated at both global and local levels.

\noindent \textbf{Selection.}
Given the plan $P$ and toolset $T$, we construct an augmented toolset $\Bar{T}$ by incorporating additional interference tools. 
Then LLMs are required to select the most appropriate tool from $\Bar{T}$ for each \textit{tool-usage steps} and generate a predicted sequence $[(s_i,p_i^{\text{TS}}), ... ,(s_j,p_j^{\text{TS}})]$, where $p_i^{\text{TS}}$ denotes the name of the chosen tool for step $s_i$.
The \textit{Key-Value based Accuracy} calculated at both global and local levels is 
the evaluator
for tool selection.

\noindent \textbf{Usage.}
Given the plan $P$, the toolset $T$, and the names of the required tools for each \textit{tool-usage step}, LLMs are tasked with generating property arguments for each \textit{tool-usage step} and outputting a predicted sequence $[(s_i,p_i^{\text{TU}}), ... ,(s_j,p_j^{\text{TU}})]$. Here, $p_i^{\text{TU}} =[\{args_j^{\text{name}}:args_j^{\text{value}}\}]_{j=1}^{k_i}$ denotes the generated arguments for step $s_i$.
The evaluation of tool usage awareness is conducted using \textit{Key-Value based Levenshtein Distance}, which is calculated at the local level. We choose this metric for the consideration that the argument values $args_j^{\text{value}}$ may be expressed in many different manners.

\begin{table*}[t]
\centering
\begin{adjustbox}{width=0.93\textwidth}
\begin{tabular}{l|cccccccccc}
\hline

\multicolumn{1}{l|}{\multirow{3}{*}{\textbf{Model}}} & \multicolumn{1}{c|}{\textbf{Planing}} & \multicolumn{3}{c|}{\textbf{Tool Creation}} & \multicolumn{5}{c|}{\textbf{Tool Usage}} & \multicolumn{1}{c}{\multirow{3}{*}{\textbf{Overall}}} \\ 
\cline{2-10}
\multicolumn{1}{l|}{} & \multicolumn{1}{c|}{\textbf{-}} & \multicolumn{2}{c}{Awareness} & \multicolumn{1}{c|}{Creation} & \multicolumn{2}{c}{Awareness} & \multicolumn{2}{c}{Selection}         & \multicolumn{1}{c|}{Usage} & \multicolumn{1}{c}{} \\ 
\cline{2-10}
\multicolumn{1}{l|}{} & \multicolumn{1}{c|}{Global} & Global & \multicolumn{1}{c}{Local} & \multicolumn{1}{c|}{Local} & \multicolumn{1}{c}{Global} & Local & \multicolumn{1}{c}{Global} & Local & \multicolumn{1}{c|}{Local} & \multicolumn{1}{c}{} \\ 
\hline
\rowcolor{red!10} \multicolumn{11}{c}{Chinese-dataset} \\ 
\hline

\multicolumn{1}{l|}{LLaMA2-7B}&\multicolumn{1}{c|}{46.44} &2.70 &6.09 &\multicolumn{1}{c|}{3.24} &2.10 &10.10 &0.30 &0.66 &3.01 & \multicolumn{1}{|c}{8.29} \\

\multicolumn{1}{l|}{ChatGLM3-6B}&\multicolumn{1}{c|}{57.54} &9.70 &23.18 &\multicolumn{1}{c|}{8.31} &12.90 &34.66  &8.50 &18.68 &29.90 & \multicolumn{1}{|c}{22.60} \\

\multicolumn{1}{l|}{Baichuan2-7B}&\multicolumn{1}{c|}{62.51} &8.20 &18.77 &\multicolumn{1}{c|}{22.39} &18.00 &46.02 &5.80 &15.48 &21.63 & \multicolumn{1}{|c}{24.31} \\

\multicolumn{1}{l|}{Vicuna-7B}&\multicolumn{1}{c|}{58.38} &7.10 &	17.77&\multicolumn{1}{c|}{22.9} &17.40 &66.49 &5.40 &11.49 &36.50 & \multicolumn{1}{|c}{27.05} \\

\multicolumn{1}{l|}{Qwen-7B}&\multicolumn{1}{c|}{61.48} &13.40 &25.91 &\multicolumn{1}{c|}{19.40} &21.40 &76.44 &12.50 &21.96 &32.35 & \multicolumn{1}{|c}{31.65} \\

\multicolumn{1}{l|}{Mistral-7B}&\multicolumn{1}{c|}{\underline{66.18}} &\underline{24.50} &\underline{42.55} &\multicolumn{1}{c|}{\underline{50.95}} &\underline{37.00} &\underline{81.30} &\underline{58.60} &\underline{74.70} &\underline{59.68} & \multicolumn{1}{|c}{\underline{55.05}} \\

\cdashline{1-11}

\multicolumn{1}{l|}{LLaMA2-13B}&\multicolumn{1}{c|}{62.05} &5.20 &16.72 &\multicolumn{1}{c|}{16.29} &28.30 &73.15 &4.10 &5.96 &25.79 & \multicolumn{1}{|c}{26.40} \\

\multicolumn{1}{l|}{Qwen-14B}&\multicolumn{1}{c|}{\underline{67.91}} &10.60 &23.90 &\multicolumn{1}{c|}{34.01} &24.90 &76.89 &13.60 &19.73 &\underline{61.10} & \multicolumn{1}{|c}{36.96} \\

\multicolumn{1}{l|}{Vicuna-13B}&\multicolumn{1}{c|}{65.72} &17.90 &37.46 &\multicolumn{1}{c|}{\underline{37.10}} &\underline{40.40} &82.56 &22.80 &39.06 &38.76 & \multicolumn{1}{|c}{42.42} \\

\multicolumn{1}{l|}{Baichuan2-13B}&\multicolumn{1}{c|}{66.84} &\underline{23.90} &\underline{45.40} &\multicolumn{1}{c|}{25.49} &34.30 &\underline{82.57} &\underline{30.80} &\underline{53.27} &59.21 & \multicolumn{1}{|c}{\underline{46.86}} \\

\cdashline{1-11}

\multicolumn{1}{l|}{LLaMA2-70B}&\multicolumn{1}{c|}{63.29} &26.90 &45.40 &\multicolumn{1}{c|}{46.03} &\underline{41.20} &81.35 &35.00 &52.21 &51.17 & \multicolumn{1}{|c}{49.17} \\

\multicolumn{1}{l|}{Qwen-72B}&\multicolumn{1}{c|}{\underline{73.40}} &\underline{36.90} &\underline{55.19} &\multicolumn{1}{c|}{\underline{61.80}} &40.80 &\underline{82.16} &\underline{72.40} &\underline{84.92} &\underline{69.52} & \multicolumn{1}{|c}{\underline{64.12}} \\

\cdashline{1-11}
\multicolumn{1}{l|}{GPT-3.5} &\multicolumn{1}{c|}{69.50} &26.20 &52.88 &\multicolumn{1}{c|}{58.00} &25.90 &79.75 &67.10 &81.50 &76.26& \multicolumn{1}{|c}{59.68} \\
\multicolumn{1}{l|}{GPT-4} &\multicolumn{1}{c|}{\textbf{76.39}} &\textbf{58.80} &\textbf{76.65} &\multicolumn{1}{c|}{\textbf{65.55} }&\textbf{60.70} &\textbf{89.76}  &\textbf{80.70} &\textbf{89.22} &\textbf{86.62}& \multicolumn{1}{|c}{\textbf{76.04}} \\
\hline
\rowcolor{blue!10}\multicolumn{11}{c}{English-dataset}\\ 
\hline

\multicolumn{1}{l|}{LLaMA2-7B}&\multicolumn{1}{c|}{65.44} &1.40 &3.74 &\multicolumn{1}{c|}{1.37} &0.40 &3.13 &0.70 &2.06 &2.14 & \multicolumn{1}{|c}{8.93} \\

\multicolumn{1}{l|}{ChatGLM3-6B}&\multicolumn{1}{c|}{60.19} &15.50 &29.69 &\multicolumn{1}{c|}{10.92} &15.20 &57.09 &13.00 &28.23 &24.84 & \multicolumn{1}{|c}{28.30} \\

\multicolumn{1}{l|}{Baichuan2-7B}&\multicolumn{1}{c|}{62.64} &10.50 &25.33 &\multicolumn{1}{c|}{17.67} &30.10 &62.24 &8.20 &17.97 &22.26 & \multicolumn{1}{|c}{28.55} \\

\multicolumn{1}{l|}{Vicuna-7B}&\multicolumn{1}{c|}{66.77} &10.10 &24.44 &\multicolumn{1}{c|}{29.54} &18.10 &61.67 &7.80 &20.43 &32.89 & \multicolumn{1}{|c}{30.19} \\

\multicolumn{1}{l|}{Qwen-7B}&\multicolumn{1}{c|}{64.81} &10.10 &24.70 &\multicolumn{1}{c|}{17.05} &22.70 &76.91 &8.80 &14.95 &32.94 & \multicolumn{1}{|c}{30.33} \\

\multicolumn{1}{l|}{Mistral-7B}&\multicolumn{1}{c|}{\underline{70.32}} &\underline{26.70} &\underline{43.34} &\multicolumn{1}{c|}{\underline{46.24}} &\underline{36.80} &\underline{77.78} &\underline{56.70} &\underline{74.44} &\underline{60.48} & \multicolumn{1}{|c}{\underline{54.76}} \\

\cdashline{1-11}

\multicolumn{1}{l|}{LLaMA2-13B}&\multicolumn{1}{c|}{68.50} &14.50 &31.00 &\multicolumn{1}{c|}{20.08} &35.70 &76.46 &16.60 &26.39 &26.41 & \multicolumn{1}{|c}{35.07} \\

\multicolumn{1}{l|}{Qwen-14B}&\multicolumn{1}{c|}{69.59} &10.50 &22.76 &\multicolumn{1}{c|}{32.21} &32.50 &\underline{79.58} &15.90 &19.55 &43.23 & \multicolumn{1}{|c}{36.20} \\

\multicolumn{1}{l|}{Vicuna-13B}&\multicolumn{1}{c|}{\underline{69.64}} &20.30 &\underline{41.50} &\multicolumn{1}{c|}{\underline{39.90}} &\underline{36.50} &79.08 &26.60 &47.22 &46.29 & \multicolumn{1}{|c}{\underline{45.23}} \\

\multicolumn{1}{l|}{Baichuan2-13B}&\multicolumn{1}{c|}{67.59} &\underline{22.30} &40.03 &\multicolumn{1}{c|}{15.79} &24.30 &78.96 &\underline{29.40} &\underline{51.34} &\underline{48.97} & \multicolumn{1}{|c}{42.08} \\

\cdashline{1-11}

\multicolumn{1}{l|}{LLaMA2-70B}&\multicolumn{1}{c|}{69.72} &25.50 &44.52 &\multicolumn{1}{c|}{46.94} &\underline{55.50} &\underline{87.61} &40.20 &53.66 &43.44 & \multicolumn{1}{|c}{51.90} \\

\multicolumn{1}{l|}{Qwen-72B}&\multicolumn{1}{c|}{\underline{72.93}} &\underline{30.90} &\underline{51.15} &\multicolumn{1}{c|}{\underline{57.98}} &49.00 &84.86 &\underline{71.80} &\underline{84.31} &\underline{63.51} & \multicolumn{1}{|c}{\underline{62.94}} \\

\cdashline{1-11}

\multicolumn{1}{l|}{GPT-3.5}&\multicolumn{1}{c|}{69.74} &33.10 &59.26 &\multicolumn{1}{c|}{51.83} &15.20 &76.29 &69.00 &82.20 &73.47& \multicolumn{1}{|c}{58.90} \\

\multicolumn{1}{l|}{GPT-4}&\multicolumn{1}{c|}{\textbf{74.07}} &\textbf{56.10} &\textbf{75.18} &\multicolumn{1}{c|}{\textbf{61.17}} &\textbf{62.50} &\textbf{90.85}  &\textbf{79.30} &\textbf{88.69} &\textbf{83.32}& \multicolumn{1}{|c}{\textbf{74.58}}\\

\hline
\end{tabular}
\end{adjustbox}
\caption{The main results of~\method. 
The overall score is the average score across all evaluation metrics.
\textbf{Bold} highlights the best score among all models, and \underline{underline} underscores the best score under the same model scale.}
\label{tab:main}
  \vspace{-0.2cm}
\end{table*}

\section{Experiments}
\label{experiments}
\subsection{Experimental Settings}
We divide all data into two parts: a test set comprising 1,000 samples for constructing evaluation datasets, and a development set with 4,824 samples, which are not used for training and can optionally serve as a development set.
For both the Chinese-dataset and English-dataset, we construct six evaluation datasets, each comprising 1,000 samples and corresponding to one of the six evaluation dimensions. 
The toolset sizes for $\hat{T}$ in tool creation awareness and $\Bar{T}$ in tool selection are both set to 8.

We evaluate closed-source and open-source LLMs on \method for both Chinese-dataset and English-dataset, 
aiming to provide comprehensive analyses for current LLMs. 
For closed-source LLMs, we select two representative models: GPT-3.5 and GPT-4 from OpenAI. The version for GPT-4 is \texttt{gpt-4-1106-preview,} and for GPT-3.5 is \texttt{gpt-3.5-turbo-1106}.
For open-source LLMs, we choose a wide spectrum of models, including: 
LLaMA2~\citep{touvron2023llama},
ChatGLM3~\citep{du2022glm}, 
Baichuan2~\citep{yang2023baichuan},
Vicuna~\citep{vicuna2023}, 
Qwen~\citep{bai2023qwen} and Mistral~\citep{jiang2023mistral}. And All experiments are run on NVIDIA V100 GPUs.

To assist LLMs in grasping the desired output format, we provide a few-shot example as a demonstration for each evaluation task. For further details, the prompts are listed in Appendix~\ref{sec:infer prompt}.

\begin{figure*}[t]
	\centering
		\centering
		\includegraphics[width=1.0\textwidth]{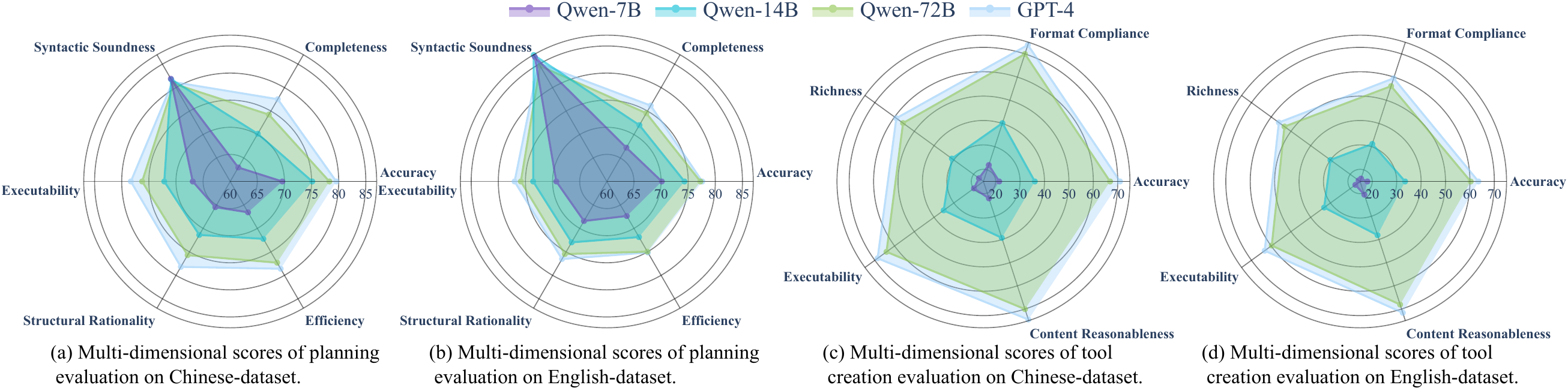}
		\caption{Multi-dimensional scores of planning and tool creation evaluation of 4 representative models under different model scales, including: Qwen-7B, Qwen-14B, Qwen-72B and GPT-4.
		 }
	\label{fig:fine_res}
   \vspace{-0.2cm}
\end{figure*}

\begin{figure*}[t]
	\centering
	\includegraphics[width=\textwidth]{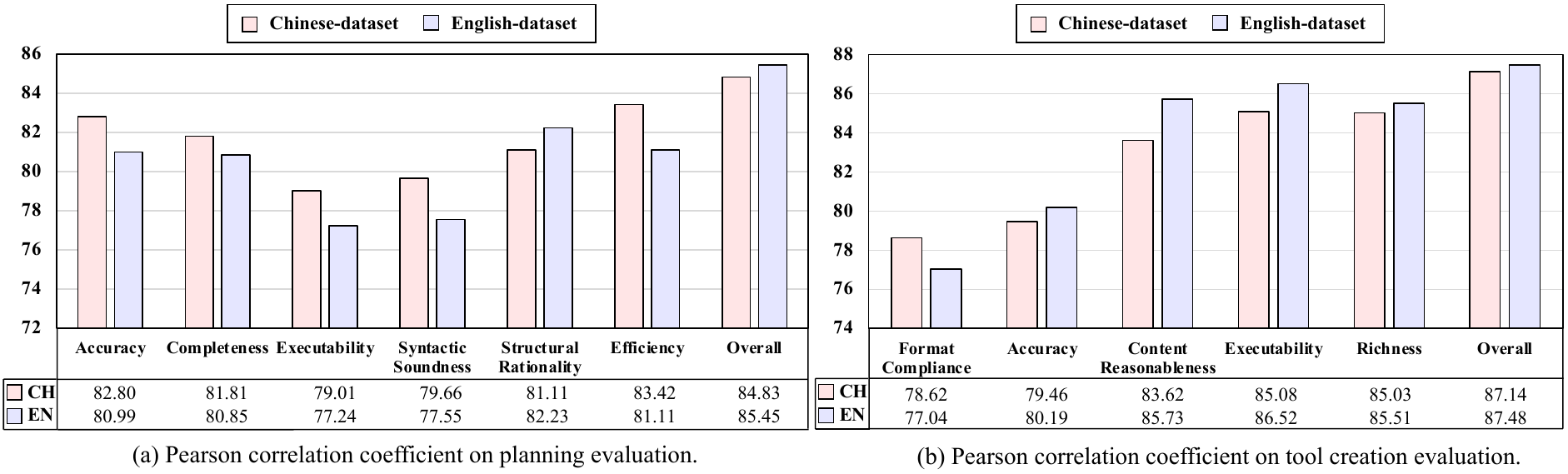}
	\caption{Pearson correlation coefficient between GPT-4 based \textit{Multi-Dimensional Point-Wise LLM-as-Judge Method} with human evaluation on planning and tool creation evaluation.
	}
	\label{fig:pearson}
   \vspace{-0.2cm}
\end{figure*}

\subsection{Main Results}
The main results are illustrated in \tablename~\ref{tab:main}. As seen, we can have the following observations:

\noindent \textbf{GPT-4 achieves the best performance.}
Among all evaluated LLMS, GPT-4 demonstrates superior performance, achieving an overall score of 76.04\% on Chinese-dataset and 74.58\% on English-dataset, 
and with particularly impressive results in the realms of tool creation and tool usage,
setting the pilot of acting as skillful tool agent. 
Besides, Mistral-7B, GPT-3.5, and Qwen-72B also get competitive performance, and Mistral-7B gets the best of two worlds between exhibiting good tool utilization abilities while keeping efficient model scale.

\noindent \textbf{The larger the model scale, the better the tool utilization ability.}
Regarding open-source LLMs, we evaluate models at three scales, approximately 7B, 13B, and 70B parameters.
The results indicate that tool utilization ability improves as the model scale increases.
This finding
aligns with prior research, which has shown that increasing of the model parameters bolsters the capabilities of LLMs~\cite{chung2022scaling, wei2022emergent}. 
While most open-source LLMs demonstrate competitive planning performance, a notable gap still persists in tool creation and usage compared to closed-source LLMs,
especially for smaller-scale models. 
This discrepancy may stem from the more intricate JSON output format requirements inherent in the evaluation dimensions of tool creation and usage. These dimensions are critical areas to focus on for enhancing LLMs' abilities in tool utilization.

\noindent \textbf{Language affects the tool utilization ability.}
Although most of LLMs support both English and Chinese, their capabilities in these two languages differ.
For open-source LLMs, in most cases, the Chinese-orientated LLMs (e.g. Qwen, Baichuan2) show better tool utilization ability on Chinese-dataset, while English-orientated LLMs (e.g. LLaMA2, Vicuna) perform better on English-dataset.
We attribute this discrepancy to variations in multilingual understanding resulting from differences in the proportions of training data.
However, despite their English orientation, GPT-3.5 and GPT-4 actually exhibit slightly weaker performance on English-dataset. We speculate that this could be due to heightened safety and alignment constraints in their English versions.

\subsection{Fine-grained Analysis of Planning and Tool Creation}
To offer a more refined analysis of the planning and tool creation capabilities of current LLMs, 
we present the multi-dimensional scores of 4 representative models under different model scales from the \textit{Multi-Dimensional Point-Wise LLM-as-Judge Method}.
More detailed multi-dimensional scores for planning and tool creation can refer to Appendix~\ref{sec:fine_plan} 
and Appendix~\ref{sec:fine_tool}, respectively.

As illustrated in \figurename~\ref{fig:fine_res} (a) and (b), we can observe that the GPT-4 and Qwen-70B are adept at breaking down complex goals into logically ordered, simpler sub-tasks. And the plans generated by smaller open-source LLMs showcase commendable \textit{Syntactic Soundness}, yet reveal evident shortcomings across the remaining five score dimensions.
This suggests that 
while many existing open-source LLMs demonstrate proficiency in generating grammatically accurate content, they still lack holistic language comprehension, particularly in fully grasping query requirements and structuring language to effectively break down tasks.

As shown in \figurename~\ref{fig:fine_res} (c) (d), it is evident that GPT-4 and Qwen-70B are significantly ahead in all score dimensions, demonstrating their proficient tool creation capabilities.
Conversely, the smaller-scale models notably trail behind them across all score dimensions, 
signaling ample room for improvement in enhancing the abilities of many open-source LLMs concerning comprehensive query understanding, adherence to prescribed output structures, and the innovation of new tools.

\subsection{Alignment of GPT-4 Scoring with Human Evaluation}\label{align}
To validate the effectiveness of the GPT-4 based \textit{Multi-Dimensional Point-Wise LLM-as-Judge Method}, we present the alignment of this metric with human evaluation on the \method dataset. Specifically, for the evaluation of planning and tool creation, we randomly selected 140 evaluation samples for human evaluation. This selection incorporates the results of all 14 evaluated LLMs to ensure diversity and mitigate bias. By providing score dimensions, inputs, and references, we request human experts to evaluate these samples and provide a multi-dimensional score, along with an overall score, following the same scoring methodology as the \textit{Multi-Dimensional Point-Wise LLM-as-Judge Method}. Subsequently, we calculate the Pearson correlation coefficient~\citep{freedman2007statistics} between human evaluation and GPT-4 evaluation for planning and tool creation. The results, depicted in \figurename~\ref{fig:pearson}, indicate a high correlation between the scores of all dimensions and the overall scores from human evaluation and GPT-4 evaluation. This suggests that the \textit{Multi-Dimensional Point-Wise LLM-as-Judge Method} is suitable for adoption in \method evaluation, as it demonstrates a strong alignment with human evaluation.

\subsection{Error Analysis}
We conduct analyses of the errors observed in evaluated LLMs, identifying five primary types:

\noindent \textbf{Not Following Instructions.}
Not following instructions is a major error type, causing a phenomenon of unanswered questions.
The Error examples 
can refer to Appendix~\ref{sec: error_follow}.

\noindent \textbf{Hallucinations.} LLMs often suffer from hallucinations~\citep{Ji_2023}, leading to outputs that include content outside the intended definition.
The error examples are listed in Appendix~\ref{sec: error_hallucination}.

\noindent \textbf{Redundant Outputs.}
Redundant outputs indicate that outputs contain unnecessary/meaningless content.
Error examples can refer to Appendix~\ref{sec: error_redundant}.

\noindent \textbf{Incomplete Outputs.}
Incomplete outputs refer to outputs that lack necessary content.
Corresponding error examples can be found in Appendix~\ref{sec: error_incomplete}.

\begin{figure}[t]
	\centering
 \includegraphics[width=0.45\textwidth]{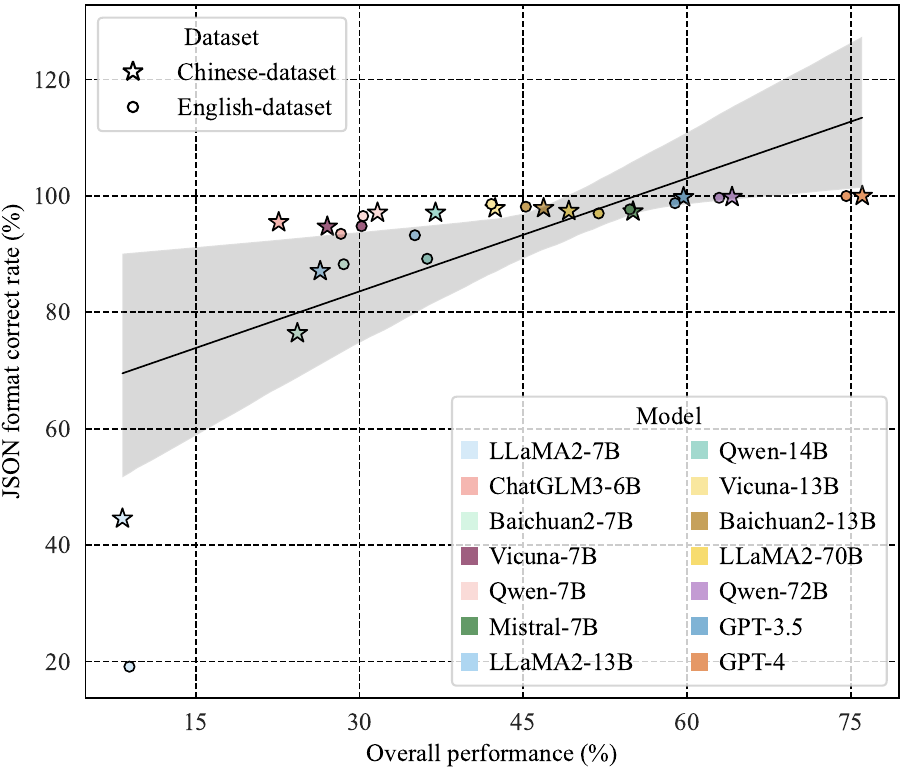}
		\caption{Correlation between JSON format correct rate with the overall score in \tablename~\ref{tab:main}.
		 }
	\label{fig:format}
   \vspace{-0.2cm}
\end{figure}

\noindent \textbf{Incorrect JSON Format:} In \method, the expected output for planning is string format, while the other five dimensions require outputs in various JSON formats. Despite thorough post-processing, errors still persist due to incorrect JSON formatting.
We calculate the proportion of outputs that adhere to the correct JSON format, namely the JSON format correct rate, and analyze its correlation with the overall performance of \method, as illustrated in Figure~\ref{fig:format}. 
It is evident that  
there exists a positive correlation between this ability and tool utilization, underscoring the importance of possessing strong skills in ensuring format compliance.
For detailed results regarding the JSON format correct rate and examples of errors caused by incorrect JSON formatting, please refer to Appendix~\ref{sec:error_format}.

\section{Related Work}\label{related-work}

\subsection{Tool Learning}
The integration of external tools enhances the capabilities of LLMs to transcend the limitations of their training data, resulting in problem-solving abilities that are not only accurate and reliable but also highly specialized \citep{qin2023tool}. The approaches to LLMs’ tool learning can broadly be divided into two categories: \textit{tool-oriented learning} and \textit{tool-augmented learning}. The former involves directly fine-tuning the LLMs to master tool usage \citep{parisi2022talm, hao2023toolkengpt, xu2023tool, shen2023hugginggpt, schick2023toolformer}, while the latter enhances LLMs with the ability to utilize tools through the provision of in-context tool descriptions and demonstrations \citep{mialon2023augmented, hsieh2023tool, patil2023gorilla, ruan2023tptu}.
There are also works that explore the tool creation of LLMs~\citep{qian-etal-2023-creator, cai2024large}, but still lack systematic evaluation of the tool creation abilities of current LLMs.
With the rapid evolution of tool learning, conducting a comprehensive evaluation that encompasses all aspects of tool utilization has become vital, which is precisely the aim of \method.

\subsection{Tool Utilization Benchmark} 
Effective benchmarks allow for 
pinpointing limitations and charting the course for future developments.
LLM tool utilization comprises three core aspects: planning, tool creation, and tool usage. Predominantly, existing benchmarks primarily focus on evaluating the model's proficiency during the tool usage phase. This particular phase involves tool usage awareness of query,  selection of the appropriate tool for each subtask \cite{huang2023metatool} and the execution of specific tools~
\cite{xu2023tool, tang2023toolalpaca, li-etal-2023-api, qin2023toolllm, ye2024tooleyes}.
Unlike prior work, we conduct a comprehensive evaluation of tool utilization, which additionally evaluates capabilities for planning and tool creation, providing a more comprehensive and fine-grained analysis of tool utilization capabilities.

\section{Conclusion}
\label{conclusion}
In this paper, we introduce \method, a comprehensive evaluation benchmark derived from real-world complex queries, aimed at evaluating the tool utilization capabilities of LLMs across six dimensions that encompass three critical aspects: planning, tool creation, and tool usage. \method excavates the necessary evaluation dimensions from actual tool utilization processes and pioneers in explicitly evaluating the NL-based planning and tool creation abilities. 
Our extensive analyses reveal that 
many current LLMs still have significant potential for enhancing their tool utilization abilities.
We hope that \method, coupled with our detailed experiments, will offer valuable insights and stimulate further research into the real-world application of LLMs in tool utilization.

\section*{Acknowledgments}
This work was partially supported by the National Natural Science Foundation of China 62176076,  Natural Science Foundation of GuangDong 2023A1515012922,
the Shenzhen Foundational Research Funding JCYJ20220818102415032, the Major Key Project of PCL2021A06, Guangdong Provincial Key Labo-ratory of Novel Security Intelligence Technologies 2022B1212010005.

\section*{Limitations}
This work contributes a comprehensive evaluation benchmark \method for tool utilization based on real-world complex queries and covering comprehensive evaluation ranges.
While promising, the tools in \method are not executable 
; rather, they are tool skeletons that represent the tools' functionalities, as many real-world demands cannot be fully addressed by existing real-world tools. 
Despite these tool skeletons are not specific implementations, they serve as simulated representations that accurately depict the tool's functionality and offer guidance for the future development of actual tools.
In the future, we plan to explore simulating the execution of our designed tools or gathering real-world data involving only executable tools.
	
\section*{Ethics Statement}
All data in \method are de-identified and safeguarding privacy concerns.
Our data construction and human evaluation processes are conducted by proficient experts, comprising postgraduate students from Chinese universities and employees of Chinese technology companies, all of whom receive fair payment for their contributions.

\bibliography{custom}
\appendix
\clearpage

\section{Data Details}

\subsection{Data Statistics}\label{sec:data_stat}
\method supports 2 languages including Chinese and English, and encompasses 22 domains, 2,032 tools, and 5,824 samples. Detailed statistics of the \method are presented in Table~\ref{tab:statistics}.
All samples within the \method involve at least one tool calling. In particular, 5,318 samples comprise multiple tool callings, whereas 506 samples consist of single tool calling, and the percentage of nested tool callings in \method is 39.61\%. On average, each sample's plan includes 12.27 steps and 2.72 tool callings, with each tool calling requiring an average of 3.05 arguments. 
And the existence of ancestral steps and \textit{tool-free steps} results in a relatively low proportion of \textit{tool-usage steps} among all steps.

\begin{table}[t]
    \centering
    \resizebox{0.48\textwidth}{!}{
    \setlength{\tabcolsep}{12pt}
    \begin{tabular}{l c}
    \toprule
         \textbf{Statistics} & \method \\
         \hline
         \# of supporting languages & 2 \\
         \# of domains & 22 \\
        \# of tools & 2,032\\
        \# of samples & 5,824\\
        -\# of single tool calling & 506\\
        -\# of multiple tool callings & 5,318\\
        percentage of nested tool callings & 39.61\%\\
    \midrule
        avg. steps per sample & 12.27 \\
        avg. tool calls per sample & 2.74 \\
        avg. arguments per tool calling & 3.05 \\
    \bottomrule
    \end{tabular}}
    \caption{Statistics of \method.}
    \label{tab:statistics}
\end{table}

\subsection{Query Collection Guideline}\label{sec:query_guideline}
\begin{figure*}[t]
	\centering
	\includegraphics[width=0.95\textwidth]{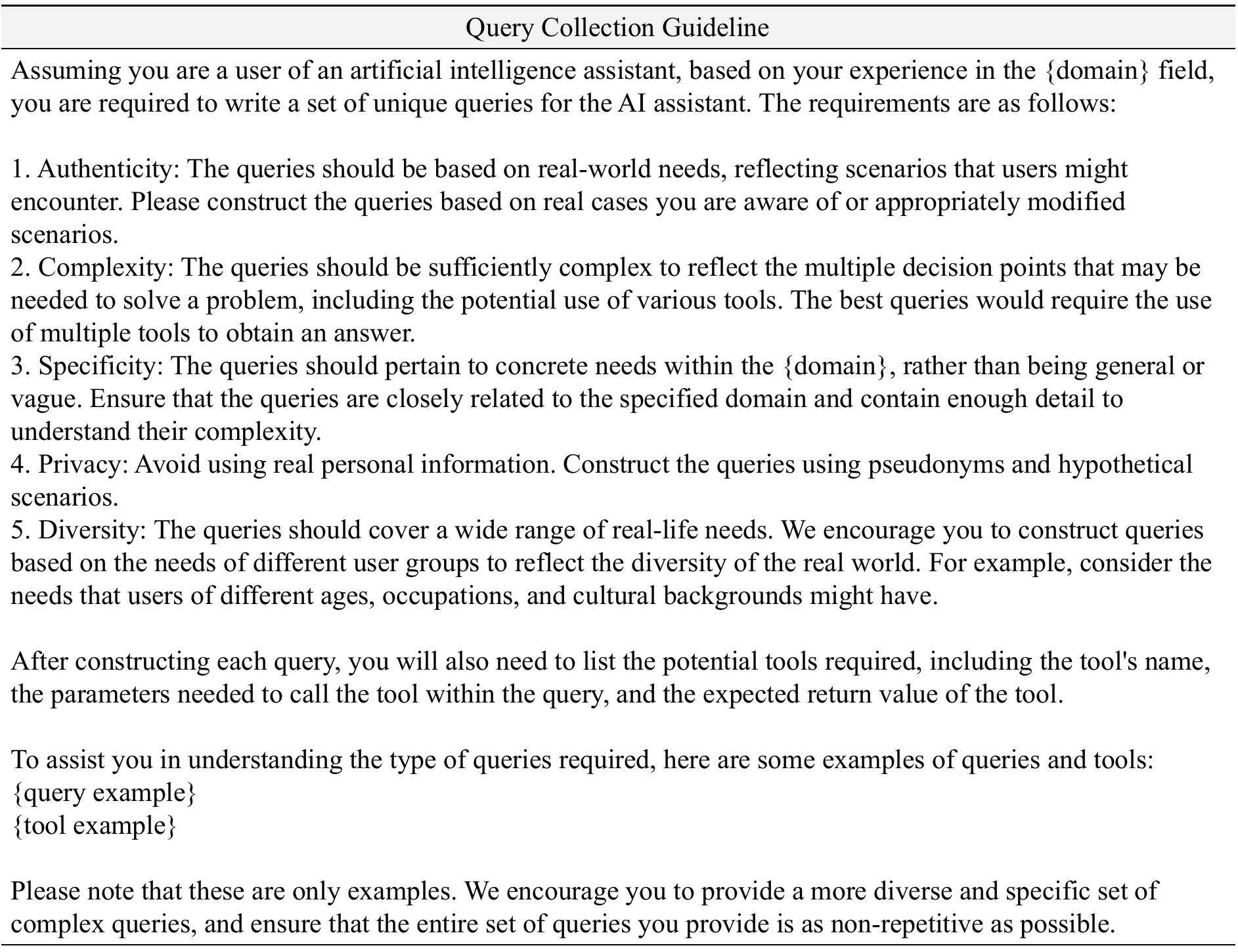}
	\caption{Guideline for query collection process.
	}
	\label{fig:query_guideline}
\end{figure*}

To ensure the high quality of collected user queries, we have developed a guideline that outlines the query collection standards for experts, which is illustrated in \figurename~\ref{fig:query_guideline}.

\subsection{Manual Refinement Guideline}\label{sec:refine_guideline}
\begin{figure*}[t]
	\centering
	\includegraphics[width=0.95\textwidth]{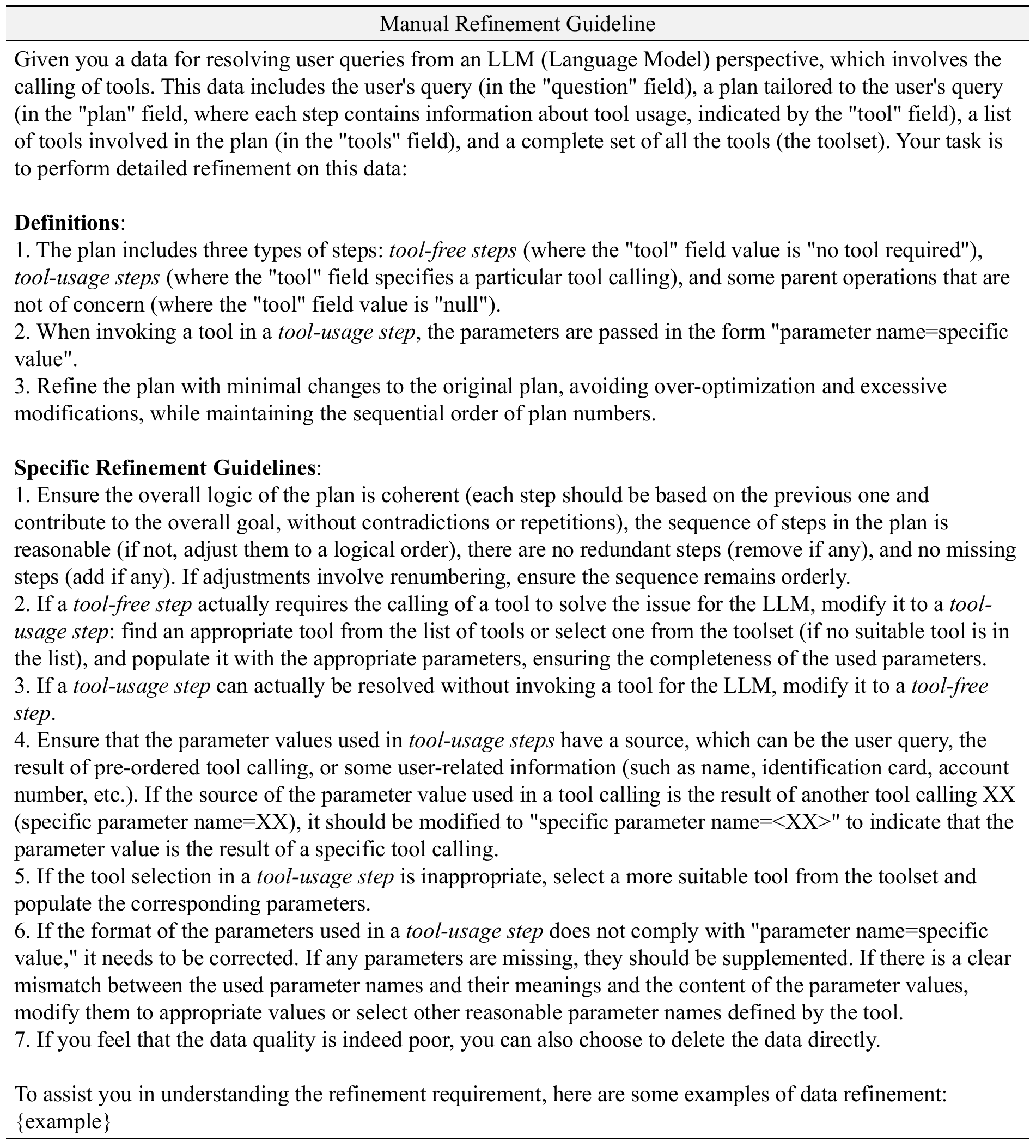}
	\caption{Guideline for manual refinement process.
	}
	\label{fig:refine_guideline}
\end{figure*}

We also provide a comprehensive manual refinement guideline for experts, as depicted in \figurename~\ref{fig:refine_guideline}.

\subsection{Generalization and Complication Examples}\label{sec:examples}
\begin{figure}[t]
	\centering
	\includegraphics[width=0.5\textwidth]{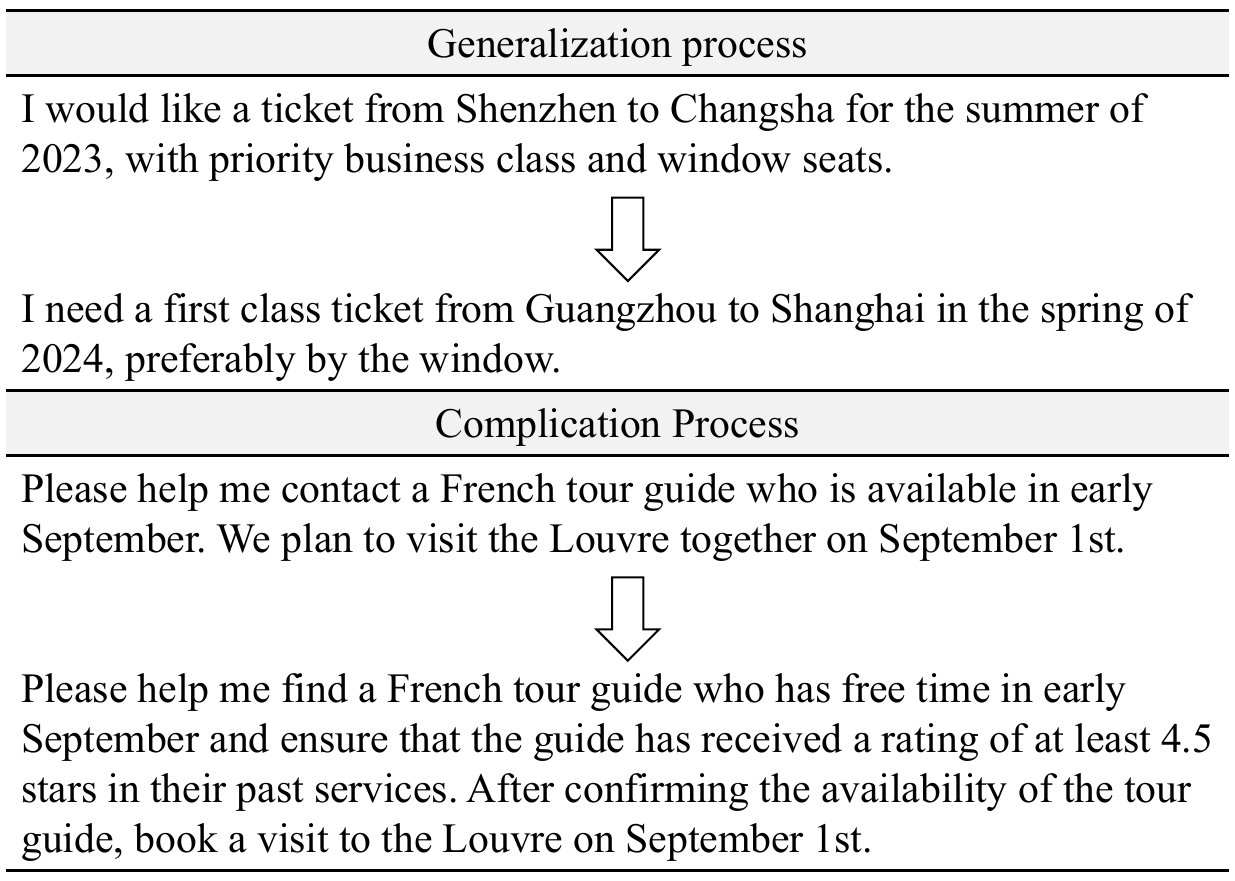}
	\caption{Examples for generalization and complication process.
	}
	\label{fig:examples}
\end{figure}

We offer examples of the generalization and complication processes, as shown in \figurename~\ref{fig:examples}.

\subsection{Prompt Template for \method Construction}\label{sec: data prompt}
\begin{figure*}[t]
	\centering
	\includegraphics[width=0.95\textwidth]{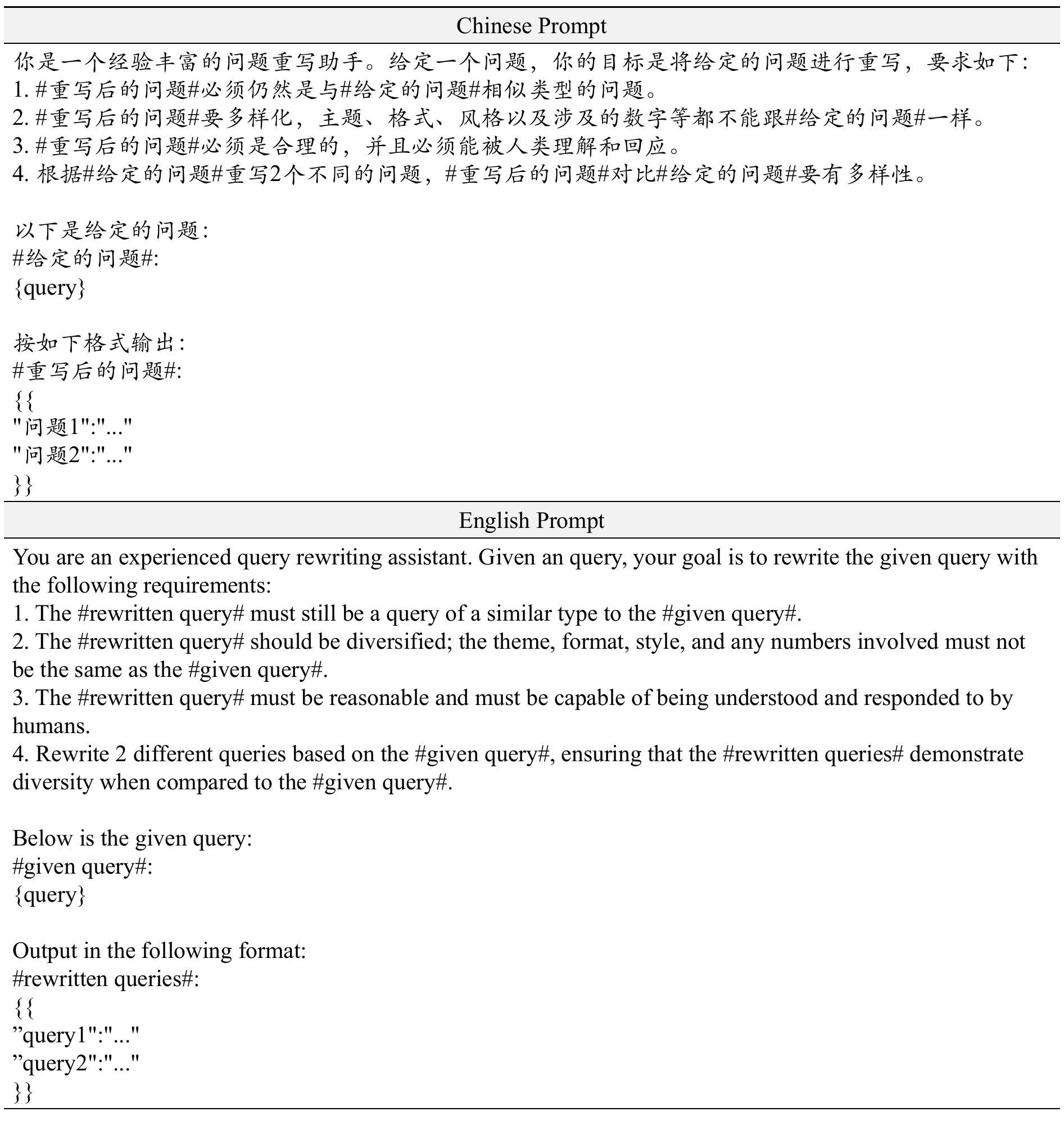}
	\caption{Prompt for query generalization process.
	}
	\label{fig:gene_prompt}
\end{figure*}

\begin{figure*}[t]
	\centering
	\includegraphics[width=0.95\textwidth]{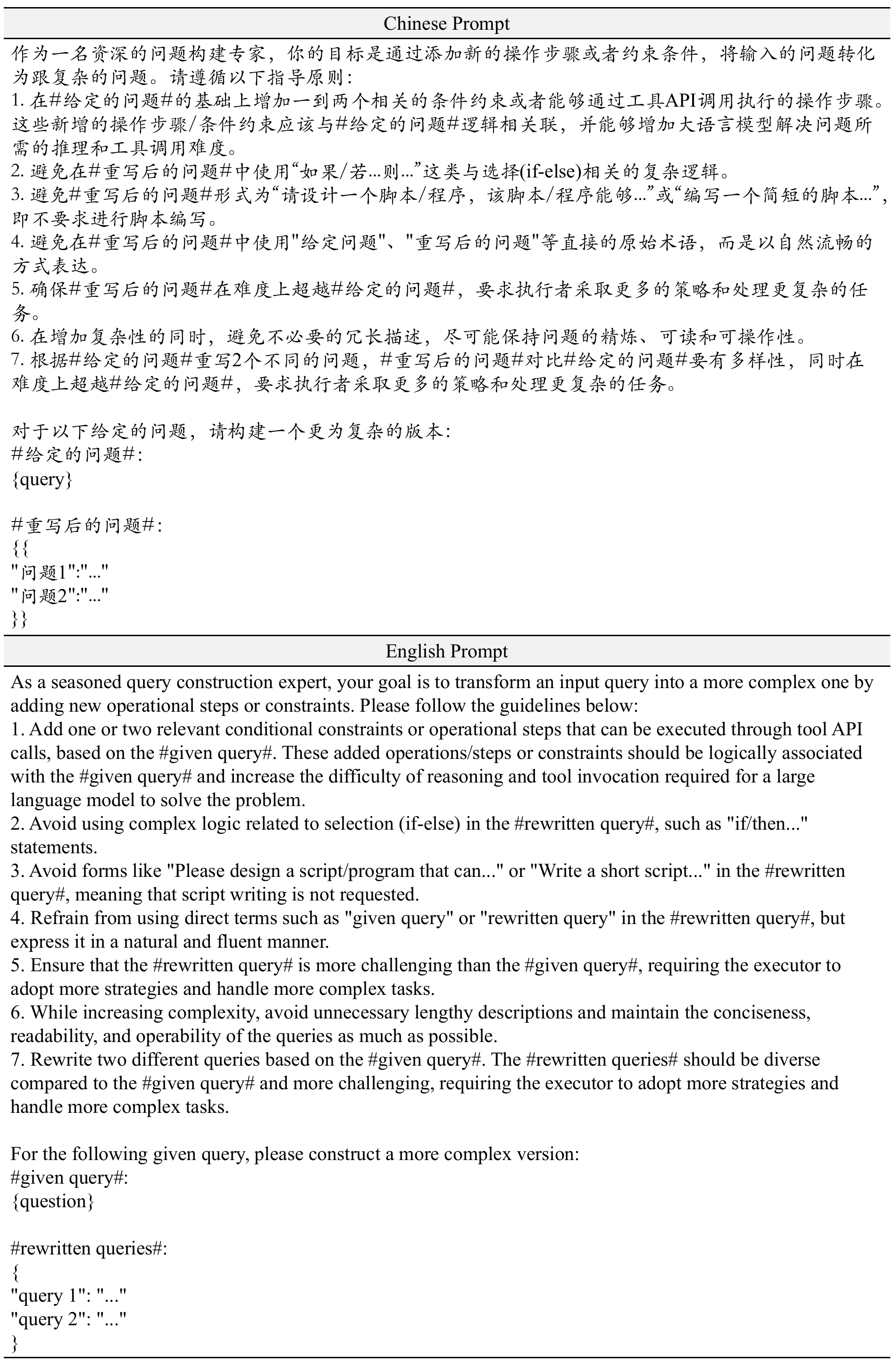}
	\caption{Prompt for query complication process.
	}
	\label{fig:evolve_prompt}
\end{figure*}

\begin{figure*}[t]
	\centering
	\includegraphics[width=0.95\textwidth]{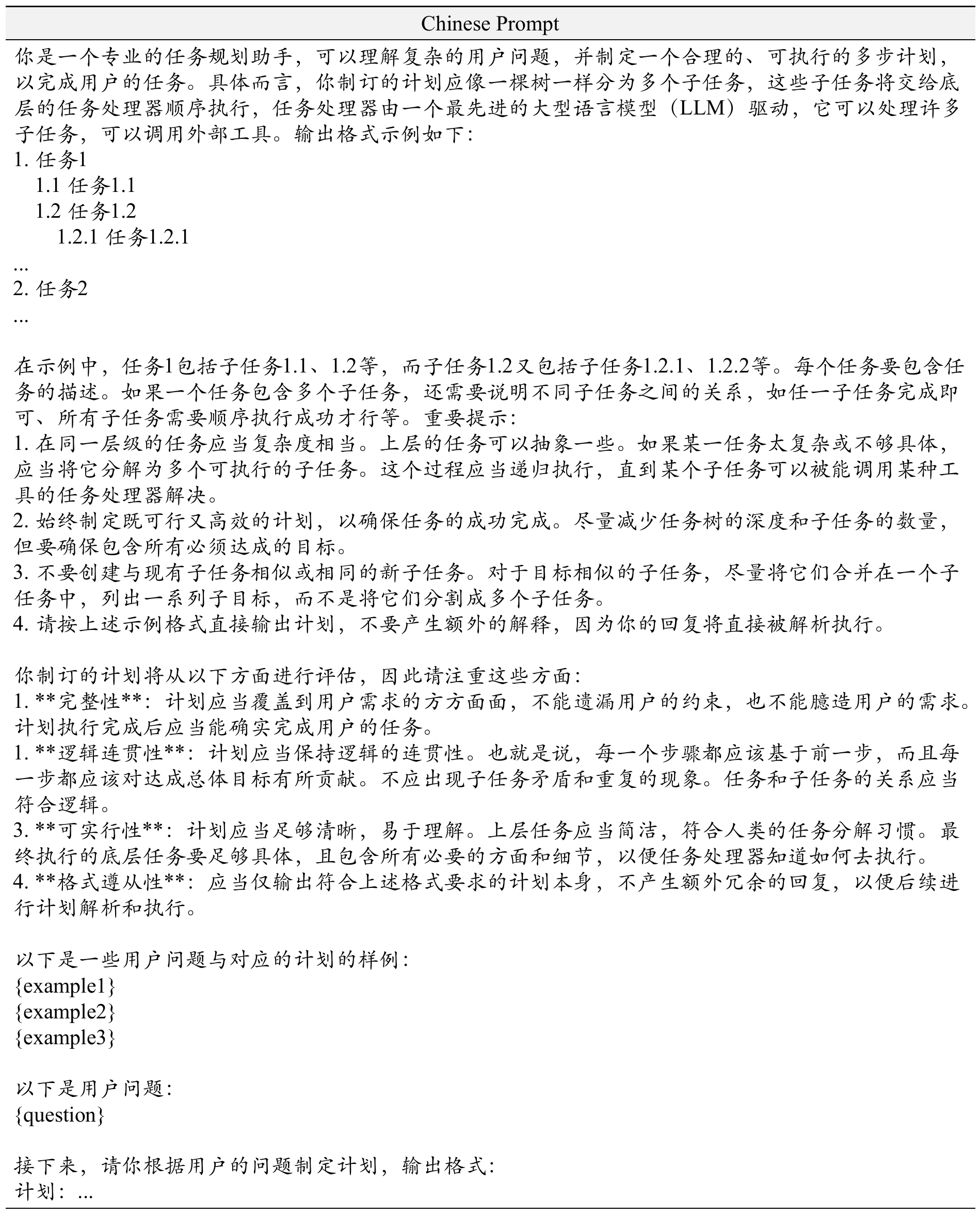}
	\caption{Chinese prompt for plan annotation process.
	}
	\label{fig:plan_prompt_ch}
\end{figure*}

\begin{figure*}[t]
	\centering
	\includegraphics[width=0.95\textwidth]{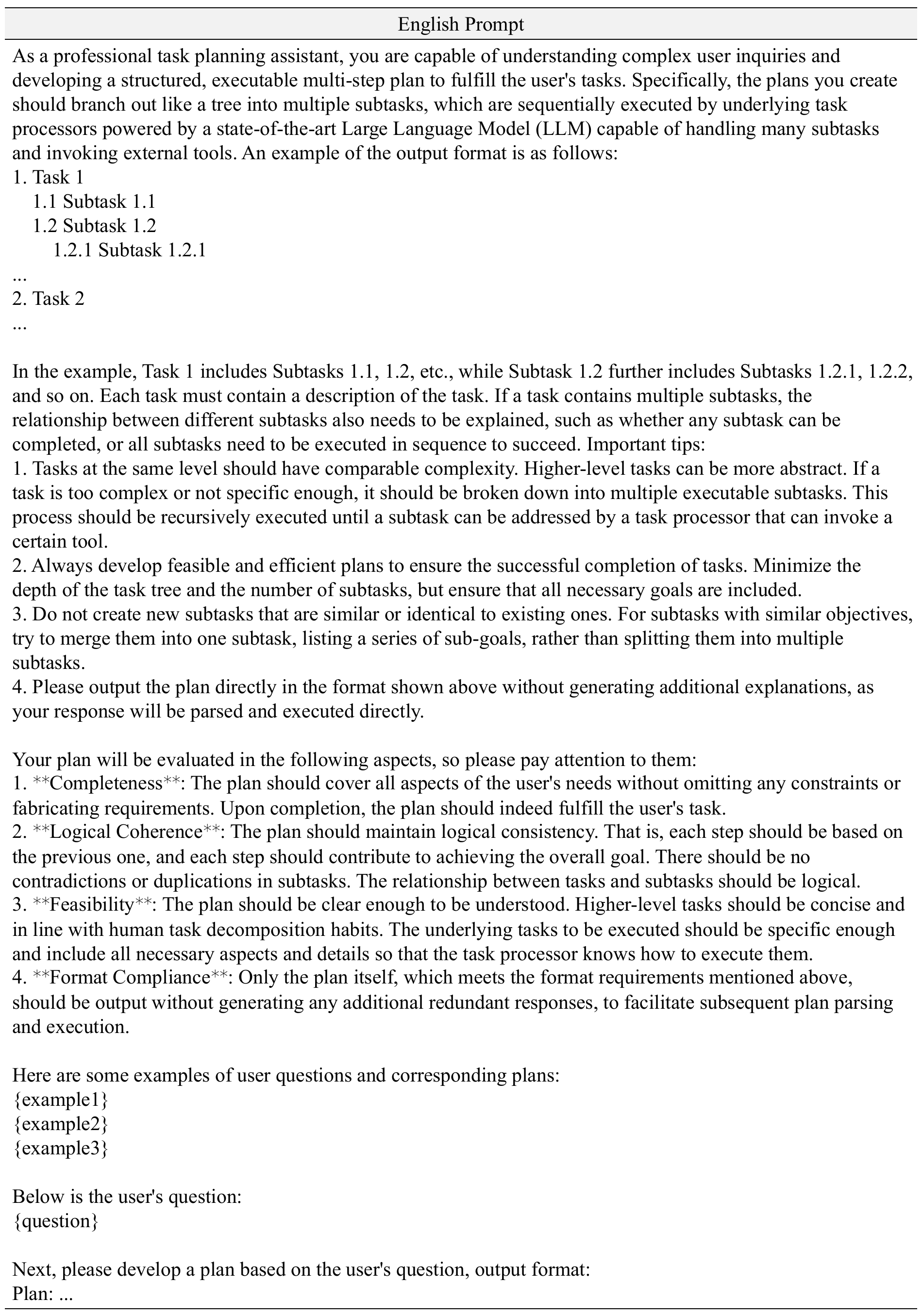}
	\caption{English prompt for plan annotation process.
	}
	\label{fig:plan_prompt_en}
\end{figure*}

\begin{figure*}[t]
	\centering
	\includegraphics[width=0.95\textwidth]{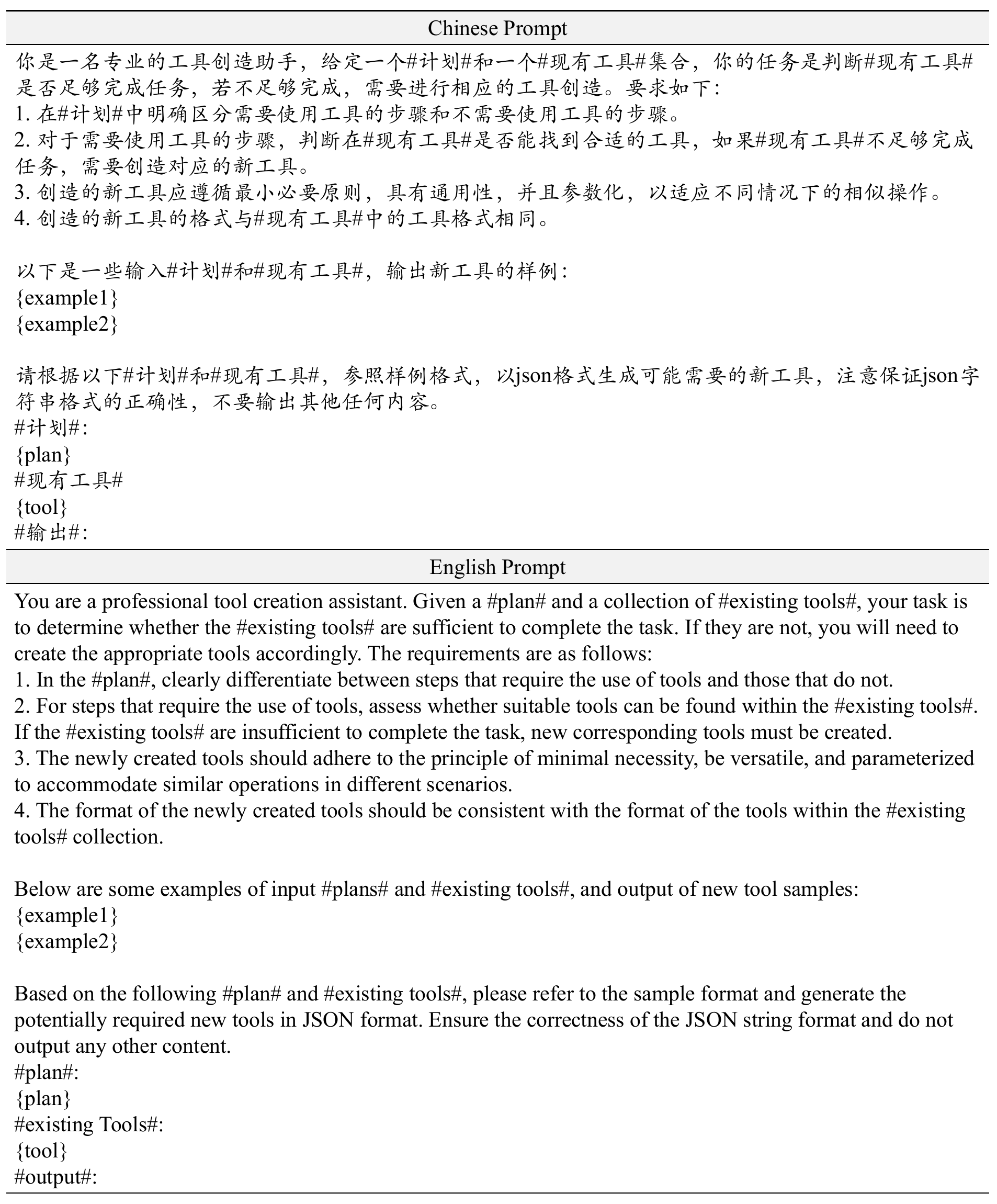}
	\caption{Prompt for tool creation process.
	}
	\label{fig:tool_create_prompt}
\end{figure*}

\begin{figure*}[t]
	\centering
	\includegraphics[width=0.95\textwidth]{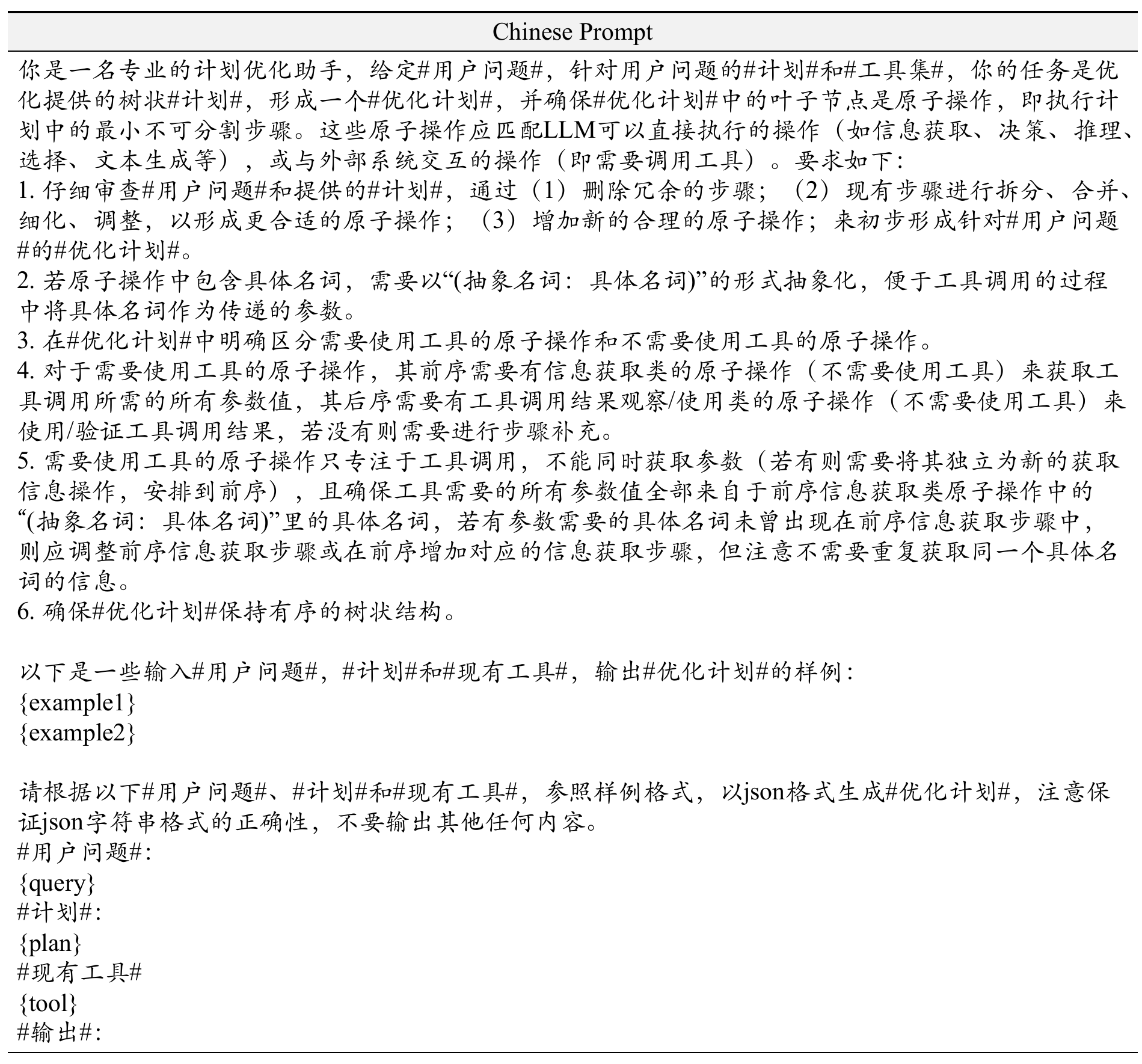}
	\caption{Chinese prompt for plan refinement process.
	}
	\label{fig:plan_refine_prompt_ch}
\end{figure*}

\begin{figure*}[t]
	\centering
	\includegraphics[width=0.95\textwidth]{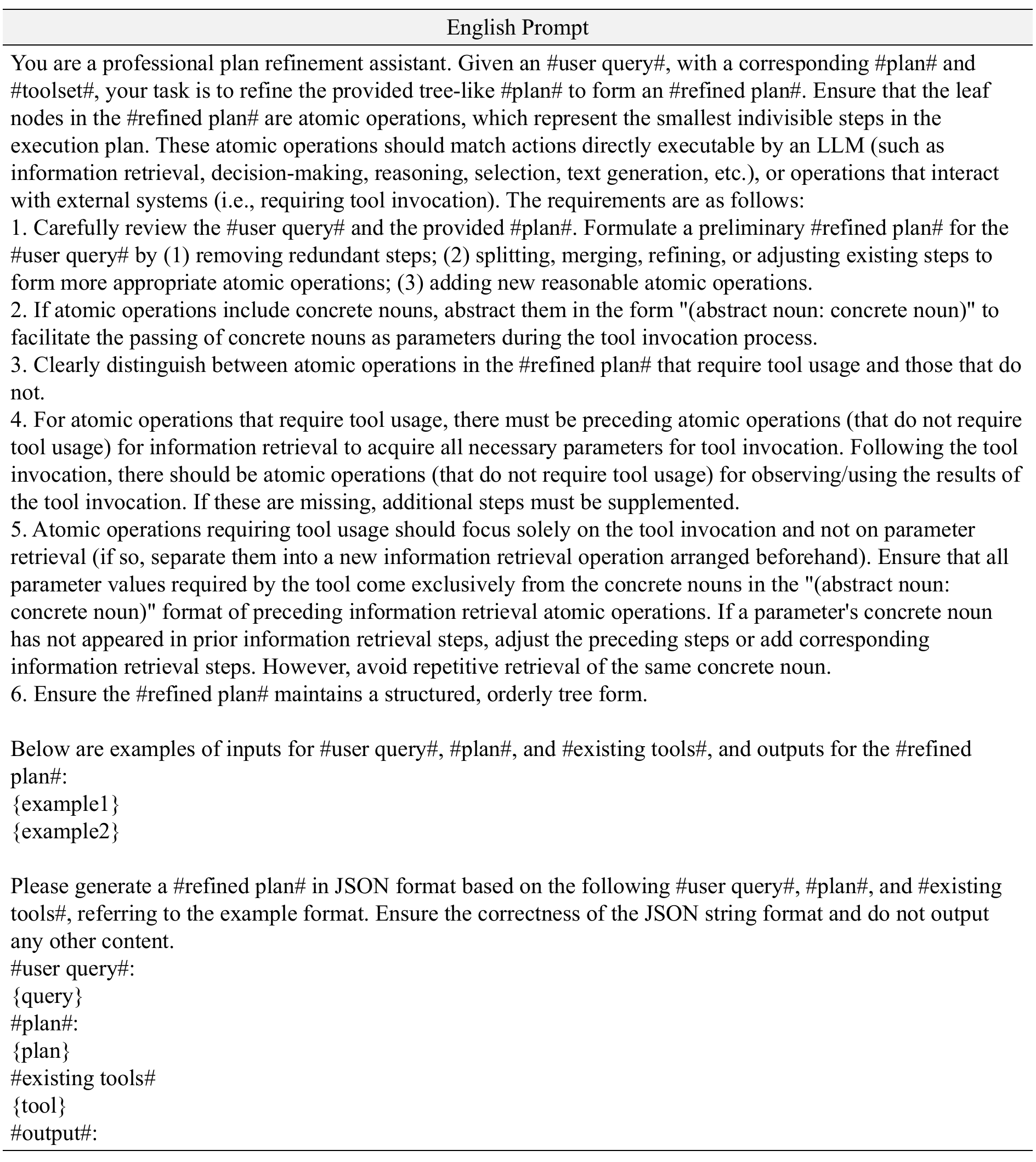}
	\caption{English prompt for plan refinement process.
	}
	\label{fig:plan_refine_prompt_en}
\end{figure*}

\begin{figure*}[t]
	\centering
	\includegraphics[width=0.92\textwidth]{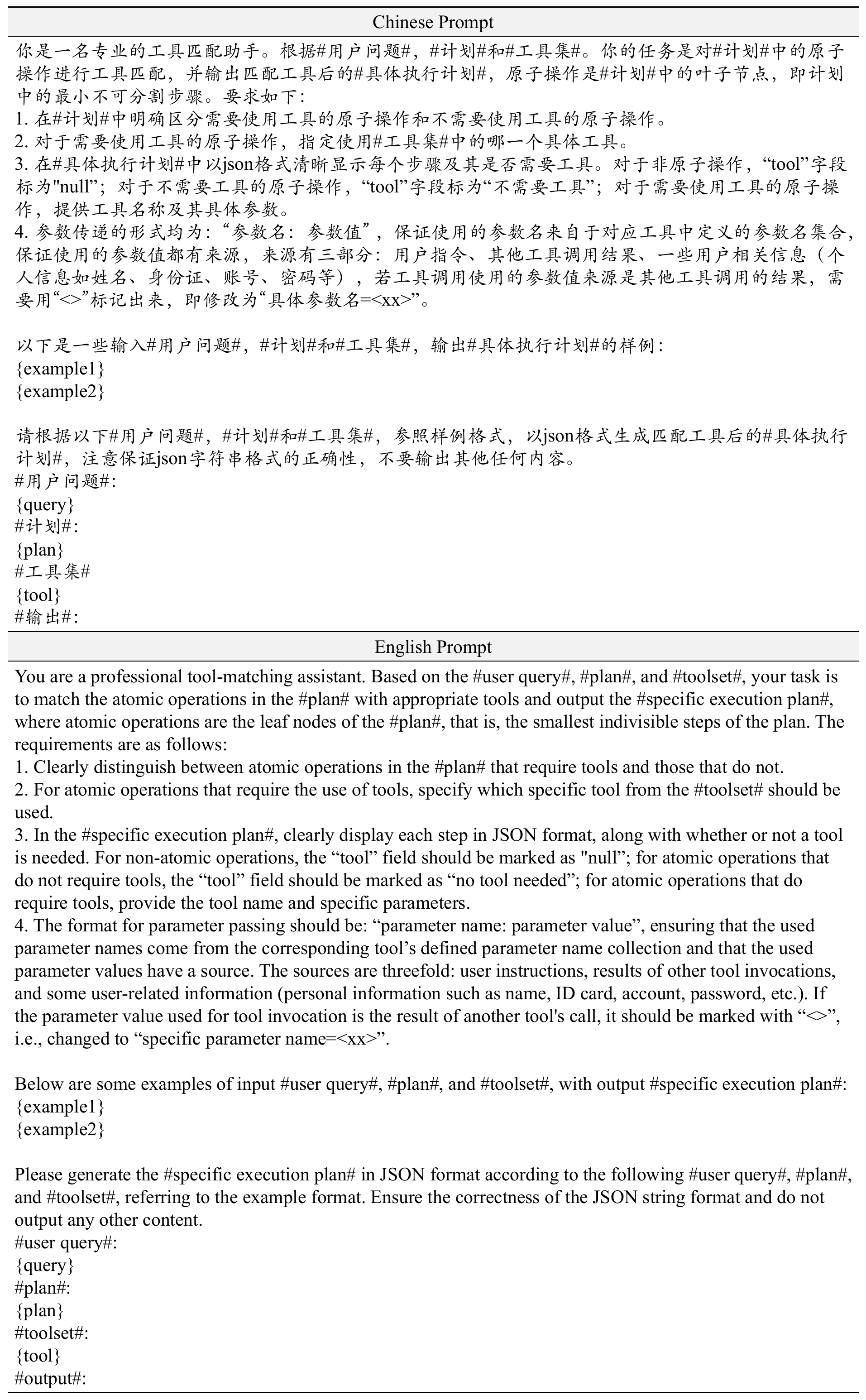}
	\caption{Prompt for tool calling message annotation process.
	}
	\label{fig:usage_prompt}
\end{figure*}

During the construction of \method, we employed GPT-4 to perform a series of automated tasks, which include query generalization (\figurename~\ref{fig:gene_prompt}), query complication (\figurename~\ref{fig:evolve_prompt}), plan annotation (\figurename~\ref{fig:plan_prompt_ch}, \figurename~\ref{fig:plan_prompt_en}), tool creation (\figurename~\ref{fig:tool_create_prompt}), plan refinement (\figurename~\ref{fig:plan_refine_prompt_ch}, \figurename~\ref{fig:plan_refine_prompt_en}), and tool calling message annotation (\figurename~\ref{fig:usage_prompt}).

\section{Evaluation Details} 

\subsection{Metric Computation Level}\label{sec:level}
Due to the multiple-step nature of the plan $P$ within \method, it is common to make several predictions $[(s_i,p_i), ... ,(s_j,p_j)]$ for the multiple steps of a given sample across distinct evaluation dimensions.
To this end, we evaluate the performance in different levels, comprising both the global level and local level, where the former considers the sample as a whole and computes the metric based on all prediction results $[p_i,...,p_j]$ in a sample, while the latter focus on individual steps and compute the metric based on a single-step prediction result $p_i$.

\subsection{Prompt Template for Multi-Dimensional Point-Wise LLM-as-Judge Method}\label{sec:eval prompt}
\begin{figure*}[t]
	\centering
	\includegraphics[width=0.95\textwidth]{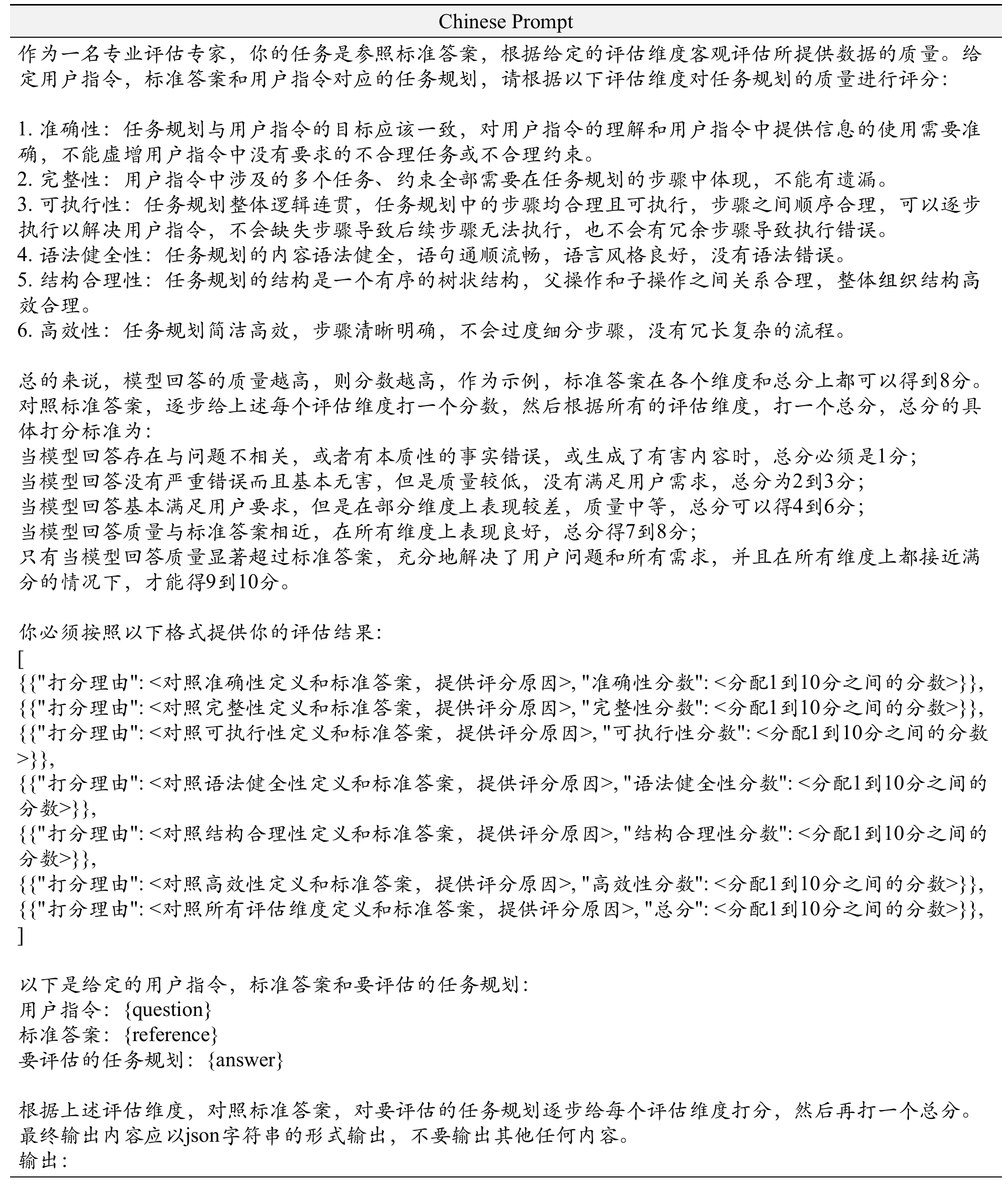}
	\caption{Prompt for \textit{Multi-Dimensional Point-Wise LLM-as-Judge Method} on Chinese-dataset planning evaluation.
	}
	\label{fig:plan_eval_ch}
\end{figure*}

\begin{figure*}[t]
	\centering
	\includegraphics[width=0.9\textwidth]{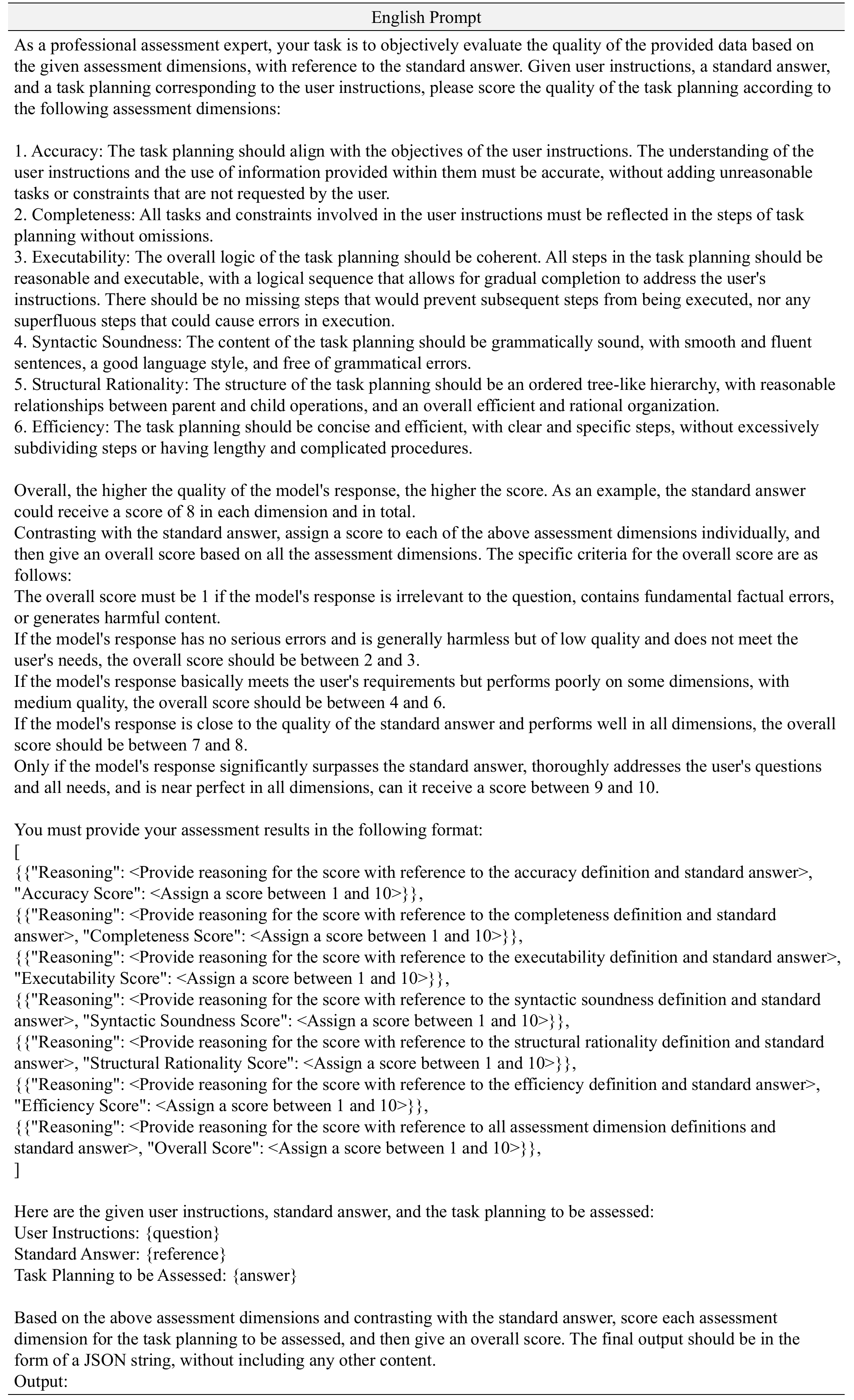}
	\caption{Prompt for \textit{Multi-Dimensional Point-Wise LLM-as-Judge Method} on English-dataset planning evaluation.
	}
	\label{fig:plan_eval_en}
\end{figure*}

\begin{figure*}[t]
	\centering
	\includegraphics[width=0.95\textwidth]{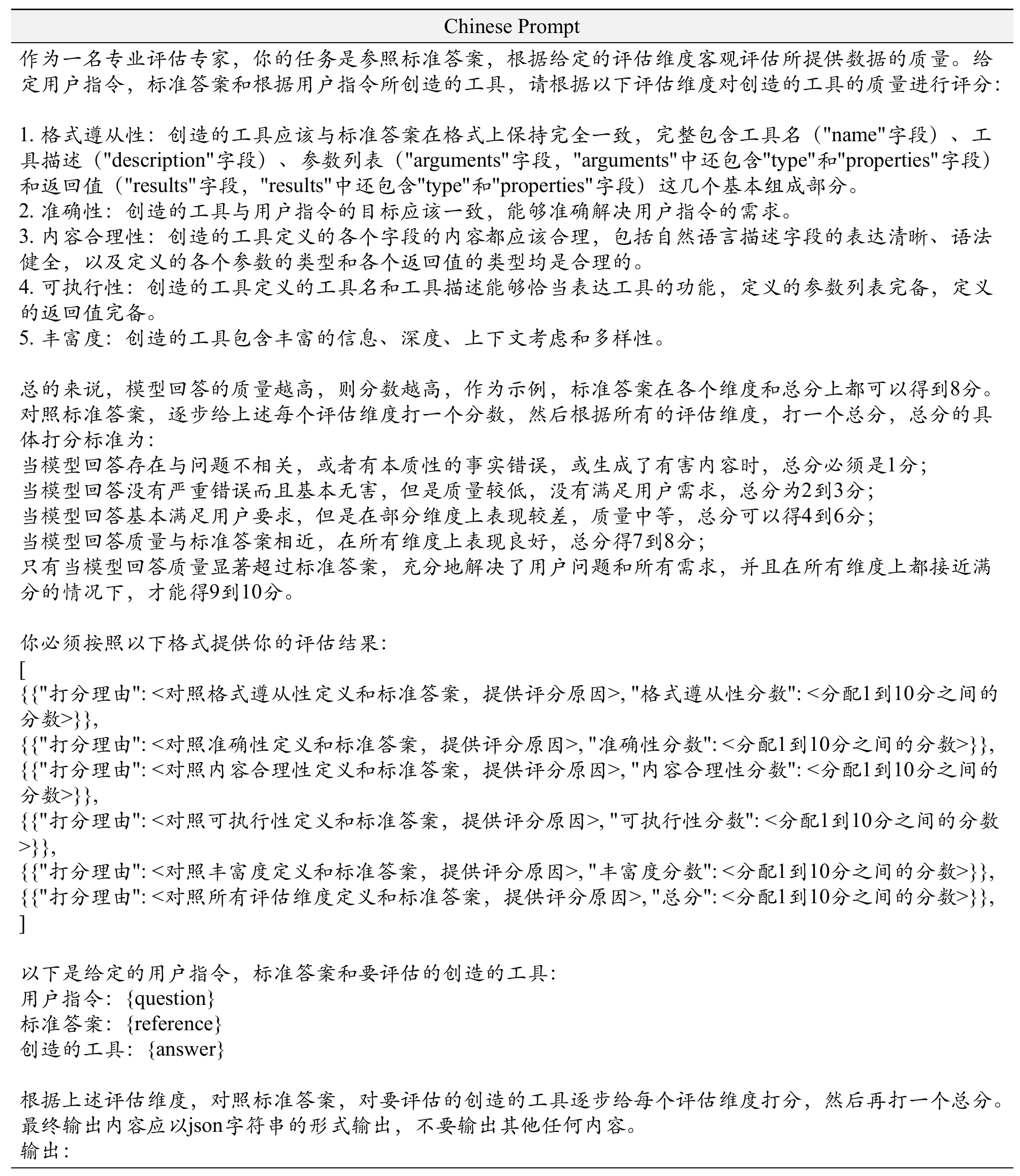}
	\caption{Prompt for \textit{Multi-Dimensional Point-Wise LLM-as-Judge Method} on Chinese-dataset tool creation evaluation.
	}
	\label{fig:tool_eval_ch}
\end{figure*}

\begin{figure*}[t]
	\centering
	\includegraphics[width=0.95\textwidth]{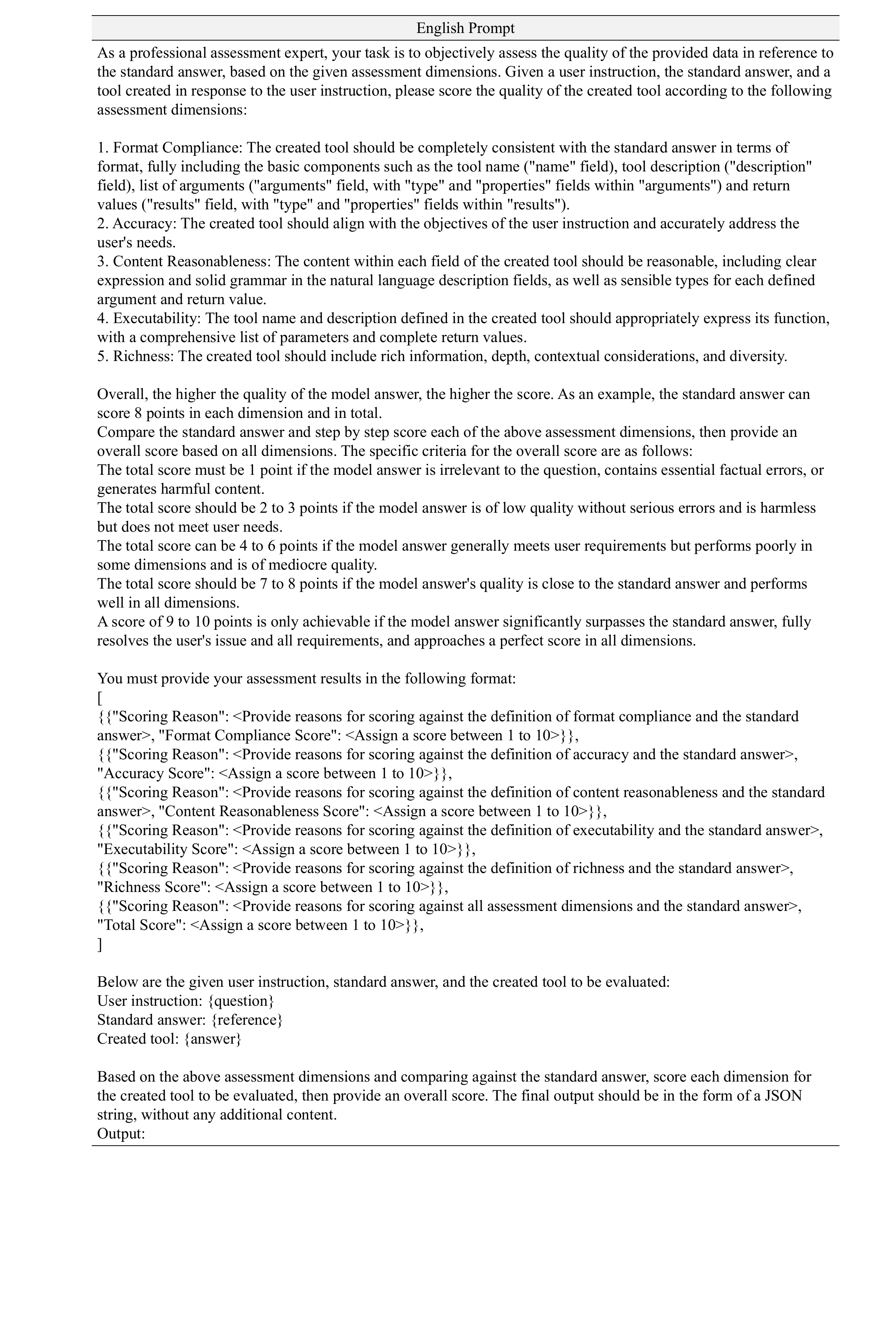}
	\caption{Prompt for \textit{Multi-Dimensional Point-Wise LLM-as-Judge Method} on English-dataset tool creation evaluation.
	}
	\label{fig:tool_eval_en}
\end{figure*}

We utilize GPT-4 to apply \textit{Multi-Dimensional Point-Wise LLM-as-Judge Method}  for planning and tool creation evaluation.
The prompt templates for Chinese-dataset and English-dataset can be found in \figurename~\ref{fig:plan_eval_ch} (planning evaluation on Chinese-dataset), \figurename~\ref{fig:plan_eval_en} (planning evaluation on English-dataset), \figurename~\ref{fig:tool_eval_ch} (tool creation evaluation on Chinese-dataset), \figurename~\ref{fig:tool_eval_en} (tool creation evaluation on English-dataset), respectively.

\begin{figure*}[t]
	\centering
	\includegraphics[width=0.95\textwidth]{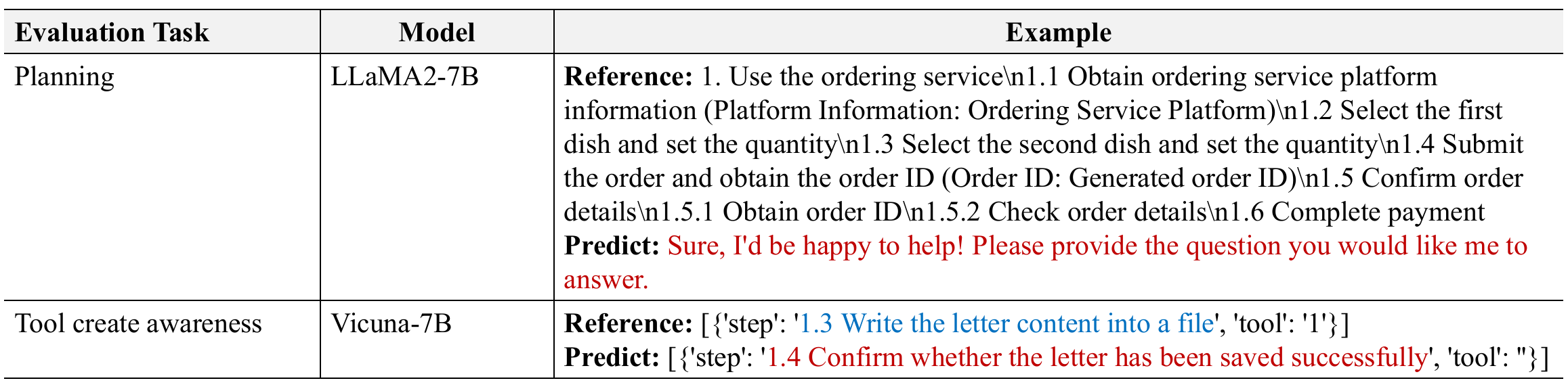}
	\caption{Error examples for not following instructions.
	}
	\label{fig:error_ins}
\end{figure*}

\subsection{Planning Evaluation Details}

\subsubsection{Score Dimension Definitions}\label{sec: plan_dimensions}
The designed six score dimensions of \textit{Multi-Dimensional Point-Wise LLM-as-Judge Method} for planning evaluation are:

\noindent (1) \textit{Accuracy:} The generated plan must be aligned with the user query's objectives.

\noindent (2) {\textit{Completeness:}} The plan should encompass all tasks and constraints mentioned in the user query, ensuring no elements are omitted.

\noindent(3) {\textit{Executability:}} Every step in the generated plan ought to be logical and executable, forming a sequence that enables the progressive fulfillment of the user's request.

\noindent(4) {\textit{Syntactic Soundness:}}  The language of the generated plan should be grammatically sound.

\noindent(5) {\textit{Structural Rationality:}} The plan should exhibit a well-organized, tree-like hierarchical structure.

\noindent(6) {\textit{Efficiency:}}  The plan must be concise and efficient, avoiding unnecessary complexity.

\subsubsection{Multi-Dimensional Scores}\label{sec:fine_plan}
We provide the detailed multi-dimensional scores of \textit{Multi-Dimensional Point-Wise LLM-as-Judge Method} on planning evaluation, as depicted in \tablename~\ref{tab:fine_res_plan}.

\begin{figure*}[t]
	\centering
	\includegraphics[width=0.95\textwidth]{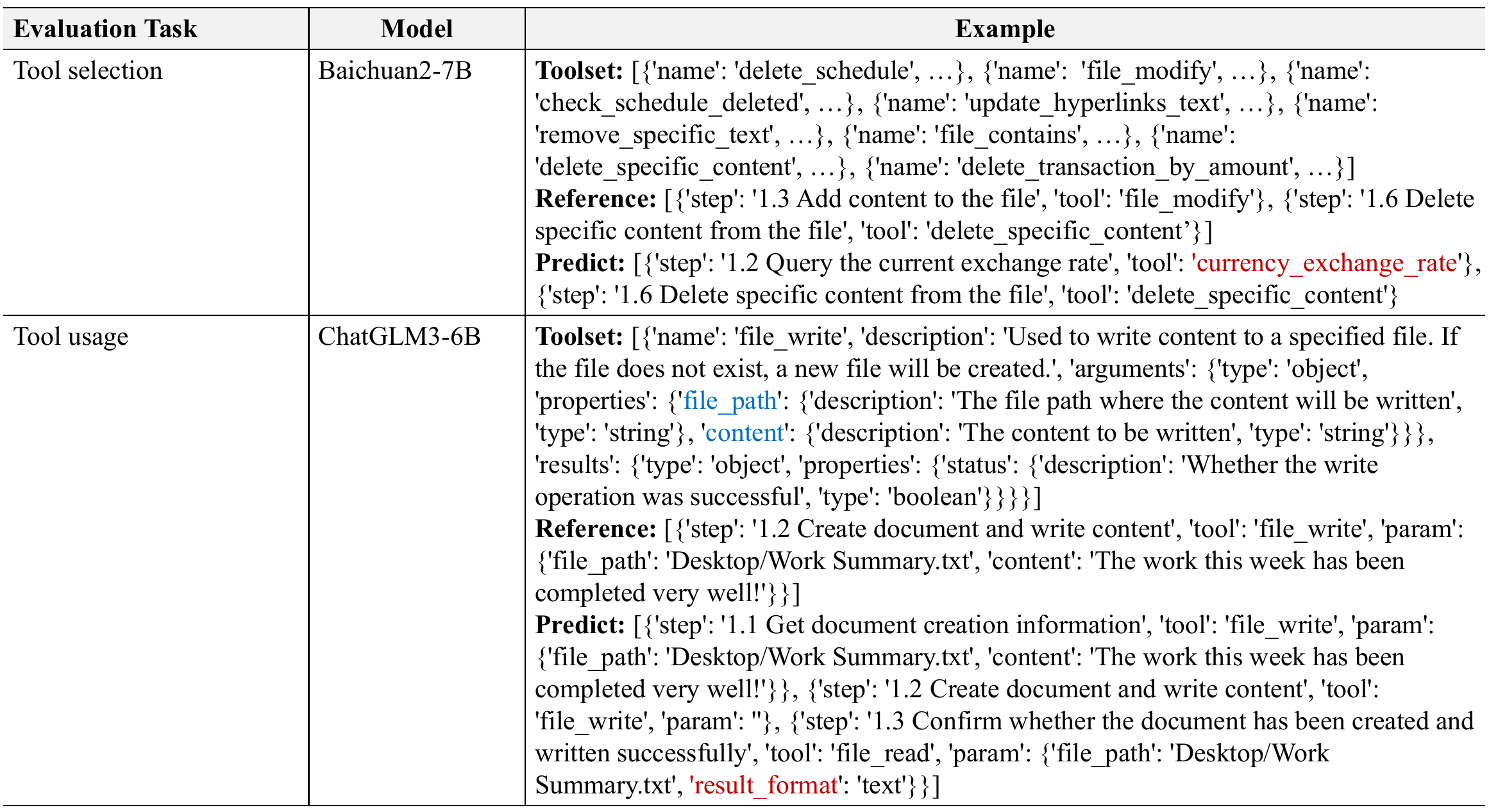}
	\caption{Error examples for hallucinations.
	}
	\label{fig:error_hallucination}
\end{figure*}

\begin{figure*}[t]
	\centering
	\includegraphics[width=0.95\textwidth]{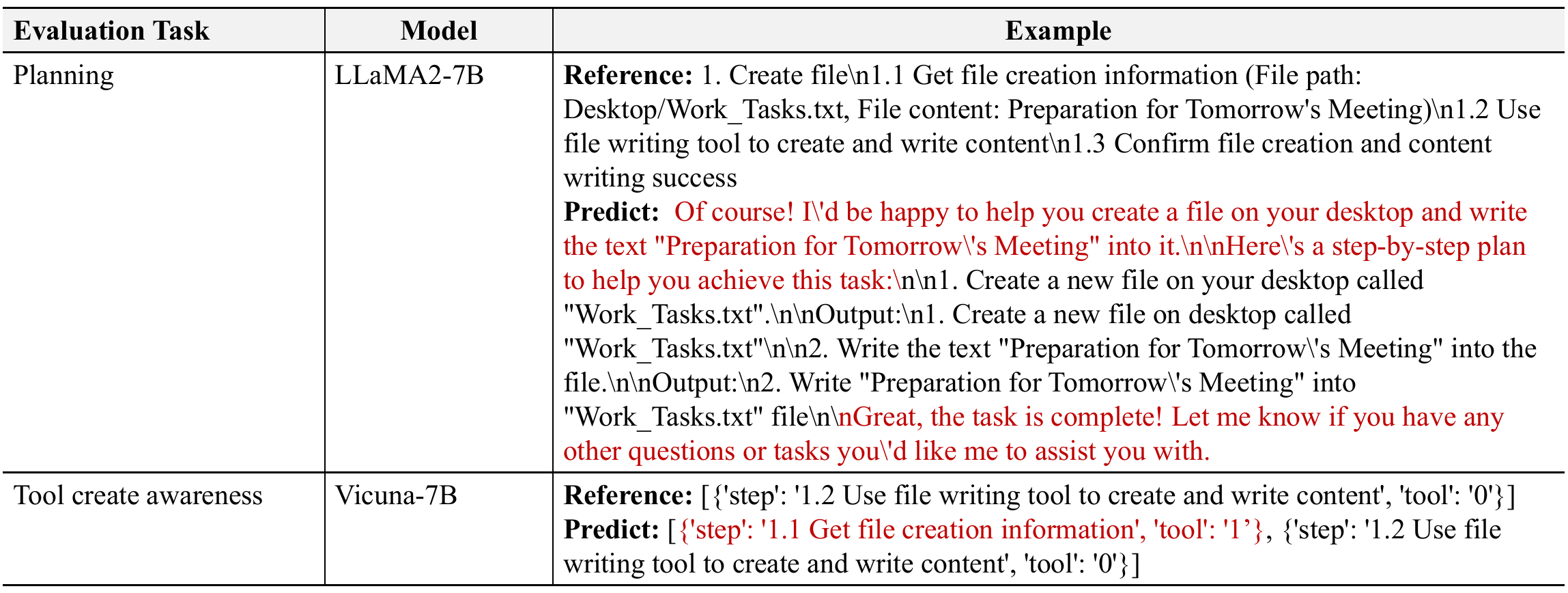}
	\caption{Error examples for redundant outputs.
	}
	\label{fig:error_redundant}
\end{figure*}

\subsection{Tool Creation Evaluation Details}

\subsubsection{Score Dimension Definitions}\label{sec: tool_dimensions}
The designed five score dimensions of \textit{Multi-Dimensional Point-Wise LLM-as-Judge Method} for tool creation evaluation are:

\noindent(1) {\textit{Format Compliance:}} The created tool must be fully consistent with the standard answer in terms of format.

\noindent(2) {\textit{Accuracy:}} The created tool must align with the objectives of the user query and accurately address the user's needs.

\noindent(3) {\textit{Content Reasonableness:}} The content within each field of the created tool should be reasonable.

\noindent(4) {\textit{Executability:}} The tool name and description in the created tool should accurately express its function, including a comprehensive list of parameters and complete return results.

\noindent(5) {\textit{Richness:}} The created tool should encompass rich information, depth, contextual considerations, and diversity.

\subsubsection{Multi-Dimensional Scores}\label{sec:fine_tool}
The detailed multi-dimensional scores of \textit{Multi-Dimensional Point-Wise LLM-as-Judge Method} on tool creation evaluation are also provided, as illustrated in \tablename~\ref{tab:fine_res_tool}.

\subsection{Error Examples}
Through comprehensive observation, we categorize errors into five primary types: (1) not following instructions; (2) hallucinations; (3) redundant outputs; (4) incomplete outputs; and (5) incorrect JSON format.

\subsubsection{Not Following Instructions}\label{sec: error_follow}
The error examples for not following instructions are shown in \figurename~\ref{fig:error_ins}. 
Such errors signify outputs that deviate from the instructions' requirements, potentially rendering them meaningless or irrelevant.

\subsubsection{Hallucinations}\label{sec: error_hallucination}
The error examples for hallucination are illustrated in \figurename~\ref{fig:error_hallucination}.
This type of error indicates that the output includes content that does not within the intended definition, such as tools hallucination for argument names hallucination.

\subsubsection{Redundant outputs}\label{sec: error_redundant}
The error examples for redundant outputs are demonstrated in \figurename~\ref{fig:error_redundant}. This type of error suggests that the output includes unnecessary or nonsensical content, such as redundant meaningless texts or redundant predictions.

\subsubsection{Incomplete outputs}\label{sec: error_incomplete}
\begin{figure*}[t]
	\centering
	\includegraphics[width=0.95\textwidth]{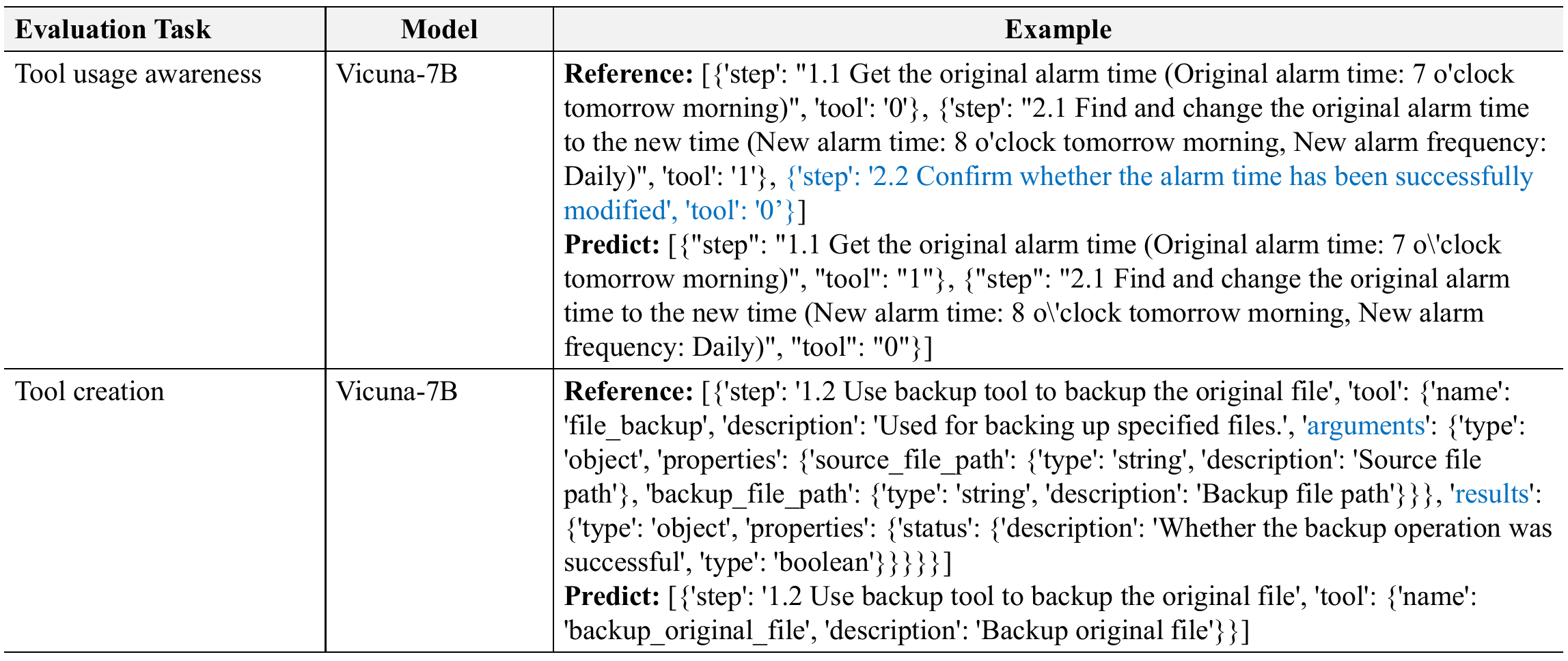}
	\caption{Error examples for incomplete outputs.
	}
	\label{fig:error_incomplete}
\end{figure*}

The error examples illustrating incomplete outputs are depicted in \figurename~\ref{fig:error_incomplete}. This type of error indicates that the output lacks essential content, such as incomplete predictions or fragmented tool structures.

\subsubsection{Incorrect JSON Format}\label{sec:error_format}
In the \method, the expected outputs for evaluation dimensions other than planning require diverse JSON formats, presenting more challenges due to the potential rendering of data unusable with even minor formatting errors.
After thorough post-processing, the proportion of outputs conforming to the correct JSON format among the evaluated LLMs, referred to as the JSON format correctness rate, is presented in \tablename~\ref{tab:format}. It is evident that 
as the model scale increases, so does the JSON format correct rate.

\begin{figure*}[t]
	\centering
	\includegraphics[width=0.95\textwidth]{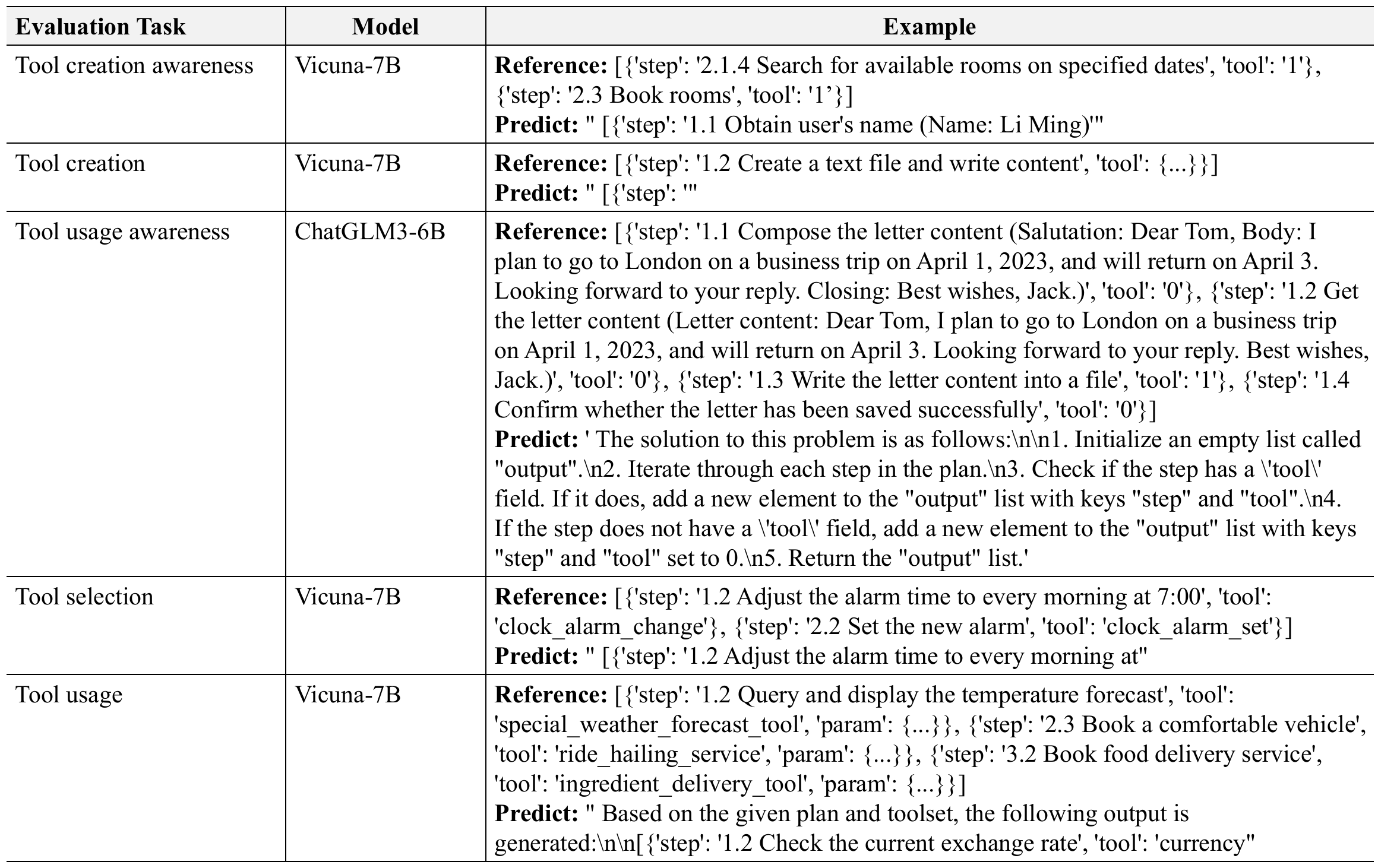}
	\caption{Error examples for incorrect JSON format.
	}
	\label{fig:error_json}
\end{figure*}

Furthermore, error examples resulting from incorrect JSON formats are depicted in \figurename~\ref{fig:error_json}. Such errors typically stem from incomplete or illegal JSON structures.

\begin{table*}[t]
\centering
\begin{adjustbox}{width=0.95\textwidth}
\begin{tabular}{l|ccccccc}
\hline
\multicolumn{1}{l|}{\multirow{2}{*}{\textbf{Model}}} & Accuracy & Completeness & Executability & Syntactic & Structural & \multicolumn{1}{c}{Efficiency} & Overall  \\ 

\multicolumn{1}{l|}{} & \multicolumn{1}{l}{} & \multicolumn{1}{l}{}  & \multicolumn{1}{c}{Soundness} & \multicolumn{1}{c}{Rationality} & \multicolumn{1}{l}{}  \\
\hline
\rowcolor{red!10} \multicolumn{8}{c}{Chinese-dataset} \\ 
\hline

\multicolumn{1}{l|}{LLaMA2-7B}& 51.35&	47.61&	50.66&	72.52&	50.79&	50.39&	46.44 \\

\multicolumn{1}{l|}{ChatGLM3-6B}& 	66.09&	59.41&	62.68&	81.47&	61.02&	62.11&	57.54\\

\multicolumn{1}{l|}{Vicuna-7B}& 65.26&	60.91&	63.51&	79.01&	63.12&	62.06&	58.38\\

\multicolumn{1}{l|}{Qwen-7B}&	69.60&	62.98&	66.90&	81.89&	65.47&	66.57&	61.48\\

\multicolumn{1}{l|}{Baichuan2-7B}&	70.16&	65.22&	67.54&	82.23&	67.04&	66.43&	62.51\\

\multicolumn{1}{l|}{Mistral-7B}&	73.00&	68.50&	70.77&	81.48&	70.07&	69.68&	66.18\\

\cdashline{1-8}
\multicolumn{1}{l|}{LLaMA2-13B}&	69.15&	63.93&	67.42&	82.12&	67.12&	66.99&	62.05\\

\multicolumn{1}{l|}{Vicuna-13B}&	73.19&	68.08&	70.64&	81.68&	69.98&	70.89&	65.72\\

\multicolumn{1}{l|}{Baichuan2-13B}&	73.67&	69.53&	71.50&	82.13&	71.14&	71.22&	66.84\\

\multicolumn{1}{l|}{Qwen-14B}&	75.13&	70.16&	72.15&	81.76&	71.39&	72.23&	67.91\\

\cdashline{1-8}
\multicolumn{1}{l|}{LLaMA2-70B}&	69.89&	66.64&	69.16&	82.42&	68.35&	66.64&	63.29\\

\multicolumn{1}{l|}{Qwen-72B}&	78.28&	74.25&	76.23&	80.83&	75.70&	77.30&	73.40\\

\cdashline{1-8}
\multicolumn{1}{l|}{GPT-3.5}&	76.43&	70.26&	73.14&	80.94&	72.26&	75.10&	69.50\\

\multicolumn{1}{l|}{GPT-4}&	79.56&	77.53&	78.31&	81.21&	78.19&	78.60&	76.39\\

\hline
\rowcolor{blue!10}\multicolumn{8}{c}{English-dataset}\\ 
\hline

\multicolumn{1}{l|}{ChatGLM3-6B}&67.07&	61.07&	64.65	&83.20	&62.45	&62.71	&60.19\\

\multicolumn{1}{l|}{Baichuan2-7B}&	67.86&	65.22&	67.58&	86.92&	67.51&	65.13&	62.64\\

\multicolumn{1}{l|}{Qwen-7B}&	70.13&	67.20&	69.31&	86.53&	68.42&	67.35&	64.81\\

\multicolumn{1}{l|}{LLaMA2-7B}& 70.57&	67.62&	69.67&	86.53&	69.77&	68.12&	65.44 \\

\multicolumn{1}{l|}{Vicuna-7B}& 71.58&	69.70&	70.85&	87.38&	70.80&	69.29&	66.77\\

\multicolumn{1}{l|}{Mistral-7B}&74.61&	72.92&	74.15&	87.23&	73.63&	72.00&	70.32\\

\cdashline{1-8}
\multicolumn{1}{l|}{Baichuan2-13B}&72.40&	70.00&	71.77&	87.47&	71.56&	69.67&	67.59\\

\multicolumn{1}{l|}{LLaMA2-13B}&72.96&	71.78&	73.25&	88.98&	73.06&	70.51&	68.50\\

\multicolumn{1}{l|}{Qwen-14B}&74.31&	72.00&	73.58&	86.95&	72.97&	71.85&	69.59\\

\multicolumn{1}{l|}{Vicuna-13B}&74.65&	72.19&	73.68&	86.14&	73.18&	71.62&	69.64\\

\cdashline{1-8}
\multicolumn{1}{l|}{LLaMA2-70B}&74.35&	72.98&	73.84&	88.20&	73.83&	70.72&	69.72\\

\multicolumn{1}{l|}{Qwen-72B}&77.24&	74.55&	75.85&	84.61&	75.47&	74.95&	72.93\\

\cdashline{1-8}
\multicolumn{1}{l|}{GPT-3.5}&76.61&	69.25&	72.13&	84.25&	71.52&	72.11&	69.74\\

\multicolumn{1}{l|}{GPT-4}&77.59&	76.17&	77.00&	85.18&	76.53&	75.14&	74.07\\

\hline
\end{tabular}
\end{adjustbox}
\caption{Multi-dimensional scores of \textit{Multi-Dimensional Point-Wise LLM-as-Judge Method} on planning evaluation.}
\label{tab:fine_res_plan}
\end{table*}

\begin{table*}[t]
\centering
\begin{adjustbox}{width=0.85\textwidth}
\begin{tabular}{l|cccccc}
\hline

\multicolumn{1}{l|}{\multirow{2}{*}{\textbf{Model}}} & Format & Accuracy & Content & Executability & \multicolumn{1}{c}{Richness} & Overall \\ 

 \multicolumn{1}{l|}{} & \multicolumn{1}{c}{Compliance} & \multicolumn{1}{l}{} &  \multicolumn{1}{c}{Reasonableness} & \multicolumn{1}{l}{} & \multicolumn{1}{l}{} \\

\hline
\rowcolor{red!10} \multicolumn{7}{c}{Chinese-dataset} \\ 
\hline

\multicolumn{1}{l|}{LLaMA2-7B}& 4.15&	3.46&	3.92&	3.39&	2.84&	3.24 \\

\multicolumn{1}{l|}{ChatGLM3-6B}&9.70&	9.10&	9.61&	8.41&	7.21&	8.31\\

\multicolumn{1}{l|}{Qwen-7B}&22.08&	21.46&	22.27&	19.80&	17.26&	19.40\\

\multicolumn{1}{l|}{Baichuan2-7B}&25.77&	24.31&	25.56&	22.93&	19.70&	22.39\\

\multicolumn{1}{l|}{Vicuna-7B}&26.83&	24.48&	26.77&	23.66&	20.41&	22.90\\

\multicolumn{1}{l|}{Mistral-7B}&58.18&	55.53&	57.84&	52.61&	45.53&	50.95\\

\cdashline{1-7}
\multicolumn{1}{l|}{LLaMA2-13B}&20.17&	17.49&	18.93&	16.77&	14.30&	16.29\\

\multicolumn{1}{l|}{Baichuan2-13B}&32.46&	27.01&	31.16&	26.43&	23.40&	25.49\\

\multicolumn{1}{l|}{Qwen-14B}&40.15&	36.04&	39.31&	35.14&	30.98&	34.01\\

\multicolumn{1}{l|}{Vicuna-13B}&42.56&	39.68&	42.43&	38.15&	33.24&	37.10\\

\cdashline{1-7}
\multicolumn{1}{l|}{LLaMA2-70B}&53.44&	48.81&	52.86&	47.42&	40.86&	46.03\\

\multicolumn{1}{l|}{Qwen-72B}&69.96&	66.86&	70.04&	64.08&	55.72&	61.80\\

\cdashline{1-7}
\multicolumn{1}{l|}{GPT-3.5}&66.19&	64.43&	64.64&	59.01&	50.30&	58.00\\

\multicolumn{1}{l|}{GPT-4}&73.87&	70.95&	74.54&	68.60&	59.11&	65.55\\

\hline
\rowcolor{blue!10}\multicolumn{7}{c}{English-dataset}\\ 
\hline

\multicolumn{1}{l|}{LLaMA2-7B}&1.33&	1.44&	1.60&	1.39&	1.17&	1.37 \\

\multicolumn{1}{l|}{ChatGLM3-6B}&10.43&	11.84&	13.04&	10.88&	9.86&	10.92\\

\multicolumn{1}{l|}{Qwen-7B}&16.24	&18.00&	20.70	&17.35&	15.74&	17.05\\

\multicolumn{1}{l|}{Baichuan2-7B}&16.97&	18.49&	20.84	&17.96&	15.91	&17.67\\

\multicolumn{1}{l|}{Vicuna-7B}&28.39	&30.68&	35.13	&30.41	&27.11	&29.54\\

\multicolumn{1}{l|}{Mistral-7B}&44.76	&48.58	&54.89&	47.64	&42.13&	46.24\\

\cdashline{1-7}
\multicolumn{1}{l|}{Baichuan2-13B}&15.49&	16.08&	19.39&	16.17&	14.64&	15.79\\

\multicolumn{1}{l|}{LLaMA2-13B}&19.26&	21.48	&24.30	&20.41	&17.76	&20.08\\

\multicolumn{1}{l|}{Qwen-14B}&31.18&	33.44&	38.15&	33.18&	30.02&	32.21\\

\multicolumn{1}{l|}{Vicuna-13B}&38.60&	41.56&	47.17	&40.85&	36.40&	39.90\\

\cdashline{1-7}
\multicolumn{1}{l|}{LLaMA2-70B}&45.28&	48.57	&55.82&	48.12&	42.16&	46.94\\

\multicolumn{1}{l|}{Qwen-72B}&56.10	&60.31&	68.07&	59.71&	53.28&	57.98\\

\cdashline{1-7}
\multicolumn{1}{l|}{GPT-3.5}&49.68&	54.91&	61.40&	52.96&	46.07&	51.83\\

\multicolumn{1}{l|}{GPT-4}&59.49&	63.47	&71.63	&63.28	&56.13	&61.17\\

\hline
\end{tabular}
\end{adjustbox}
\caption{Multi-dimensional scores of \textit{Multi-Dimensional Point-Wise LLM-as-Judge Method} on tool creation evaluation.}
\label{tab:fine_res_tool}
\end{table*}

\begin{table*}[t]
\centering
\begin{adjustbox}{width=0.75\textwidth}
\begin{tabular}{l|cccccc}
\hline

\multicolumn{1}{l|}{\multirow{2}{*}{\textbf{Model}}}& \multicolumn{2}{c|}{\textbf{Tool Creation}} & \multicolumn{3}{c|}{\textbf{Tool Usage}} &\multicolumn{1}{l} {\multirow{2}{*}{\textbf{Avg.}}} \\ 
\cline{2-6}
\multicolumn{1}{l|}{} & \multicolumn{1}{c}{Awareness} & \multicolumn{1}{c|}{Creation} & \multicolumn{1}{c}{Awareness} & \multicolumn{1}{c}{Selection}         & \multicolumn{1}{c|}{Usage} & \multicolumn{1}{c}{} \\ 
\hline
\rowcolor{red!10} \multicolumn{7}{c}{Chinese-dataset} \\ 
\hline

\multicolumn{1}{l|}{LLama2-7B}&50.70&	54.30&	41.00&	35.70&	41.10&	44.56 \\

\multicolumn{1}{l|}{Baichuan2-7B}&78.50&	70.80&	70.00&	83.80&	79.00&	76.42\\

\multicolumn{1}{l|}{Vicuna-7B}&92.10&	89.40&	99.10&	98.20&	94.80&	94.72\\

\multicolumn{1}{l|}{ChatGLM3-6B}&97.80&	86.80&	96.00&	98.80&	98.00&	95.48\\

\multicolumn{1}{l|}{Qwen-7B}&97.10&	92.90&	99.70&	98.20&	97.70&	97.12\\

\multicolumn{1}{l|}{Mistral-7B}&99.20&	89.00&	99.60&	99.50&	99.30&	97.32\\

\cdashline{1-7}
\multicolumn{1}{l|}{LLama2-13B}&88.00&	88.00&	99.70&	73.90&	85.80&	87.08\\

\multicolumn{1}{l|}{Qwen-14B}&97.10&	92.90&	99.70&	98.20&	97.70&	97.12\\

\multicolumn{1}{l|}{Vicuna-13B}&99.80&	94.40&	99.80&	98.30&	97.10&	97.88\\

\multicolumn{1}{l|}{Baichuan2-13B}&99.80&	90.60&	99.90&	100.00&	99.20&	97.90\\

\cdashline{1-7}
\multicolumn{1}{l|}{LLama2-70B}&99.70&	91.30&	100.00&	98.10&	98.00&	97.42\\

\multicolumn{1}{l|}{Qwen-72B}&100.00&	99.30&	99.90&	100.00&	99.90&	99.82\\

\cdashline{1-7}
\multicolumn{1}{l|}{GPT-3.5}&100.00&	99.00&	100.00&	100.00&	99.90&	99.78\\

\multicolumn{1}{l|}{GPT-4}&100.00&	100.00&	100.00&	100.00&	100.00&	100.00\\

\hline
\rowcolor{blue!10}\multicolumn{7}{c}{English-dataset}\\ 
\hline

\multicolumn{1}{l|}{LLama2-7B}&29.40&	10.50&	3.60&	31.10&	21.00&	19.12 \\

\multicolumn{1}{l|}{Baichuan2-7B}&95.50&	81.00&	80.80&	98.50&	85.50&	88.26\\

\multicolumn{1}{l|}{ChatGLM3-6B}&91.40&	86.70&	96.40&	97.10&	95.80&	93.48\\

\multicolumn{1}{l|}{Vicuna-7B}&96.90&	92.40&	99.40&	98.10&	87.20&	94.80\\

\multicolumn{1}{l|}{Qwen-7B}&97.70&	91.80&	99.70&	97.40&	96.10&	96.54\\

\multicolumn{1}{l|}{Mistral-7B}&99.40&	90.00&	99.90&	99.50&	99.60&	97.68\\

\cdashline{1-7}
\multicolumn{1}{l|}{Qwen-14B}&89.90&	94.70&	99.50&	67.00&	94.90&	89.20\\

\multicolumn{1}{l|}{LLama2-13B}&97.10&	89.80&	99.80&	91.50&	88.00&	93.24\\

\multicolumn{1}{l|}{Vicuna-13B}&99.90&	93.30&	99.60&	100.00&	97.90&	98.14\\

\multicolumn{1}{l|}{Baichuan2-13B}&99.60&	95.10&	99.80&	100.00&	98.50&	98.60\\

\cdashline{1-7}
\multicolumn{1}{l|}{LLama2-70B}&99.00&	93.50&	99.90&	94.90&	97.70&	97.00\\

\multicolumn{1}{l|}{Qwen-72B}&100.00&	99.00&	99.60&	100.00&	99.80&	99.68\\

\cdashline{1-7}
\multicolumn{1}{l|}{GPT-3.5}&100.00&	94.00&	100.00&	99.80&	100.00&	98.76\\

\multicolumn{1}{l|}{GPT-4}&100.00&	100.00&	99.90&	100.00&	100.00&	99.98\\

\hline
\end{tabular}
\end{adjustbox}
\caption{The results of JSON format correct rate for tasks that require JSON output format. And AVG. is the average score of all five tasks' JSON format correct rate.}
\label{tab:format}
\end{table*}

\subsection{Prompt Template for Inference}\label{sec:infer prompt}

\begin{figure*}[t]
	\centering
	\includegraphics[width=0.95\textwidth]{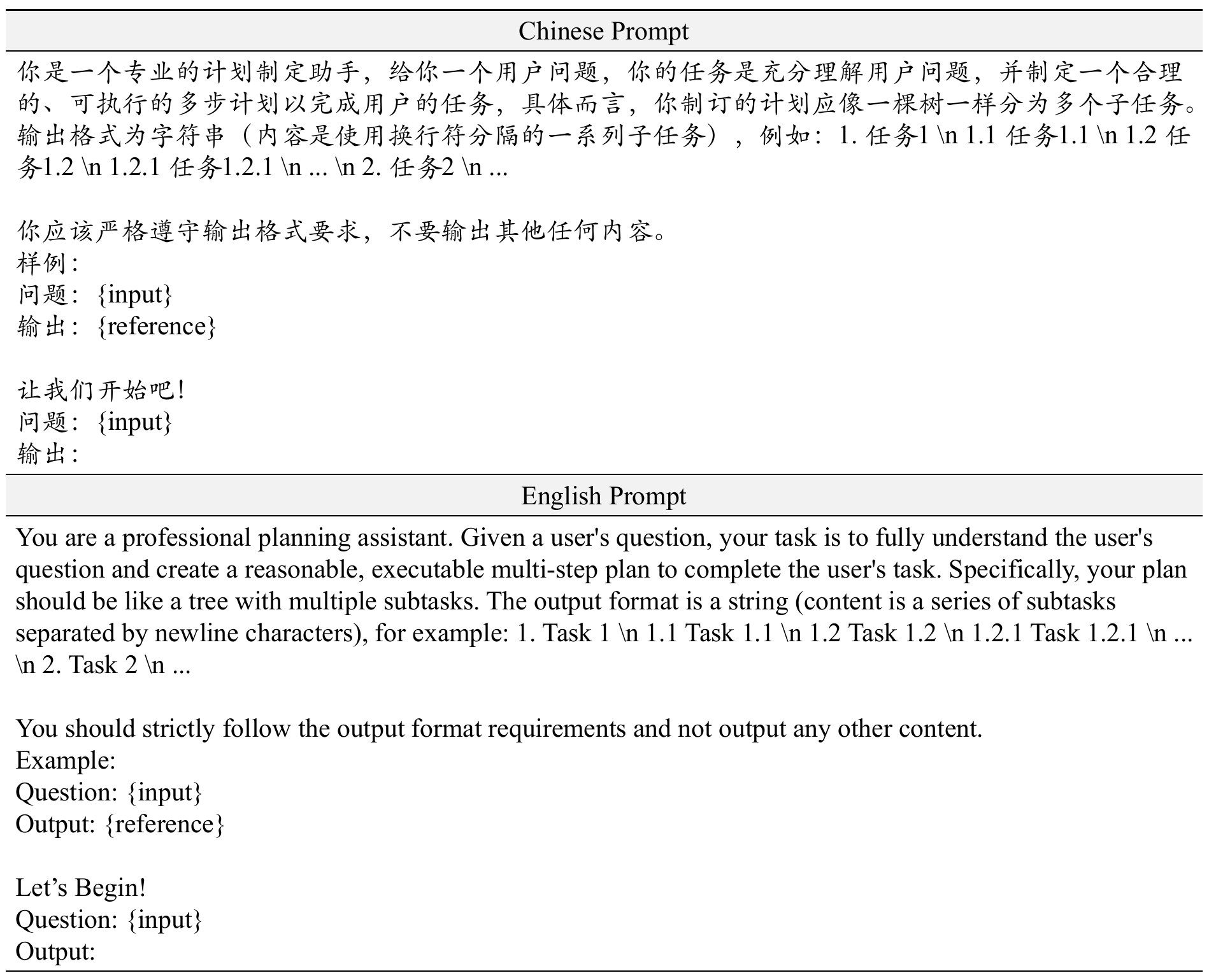}
	\caption{Prompt for planning inference on Chinese-dataset and English-dataset.
	}
	\label{fig:planning_prompt}
\end{figure*}

\begin{figure*}[t]
	\centering
	\includegraphics[width=0.95\textwidth]{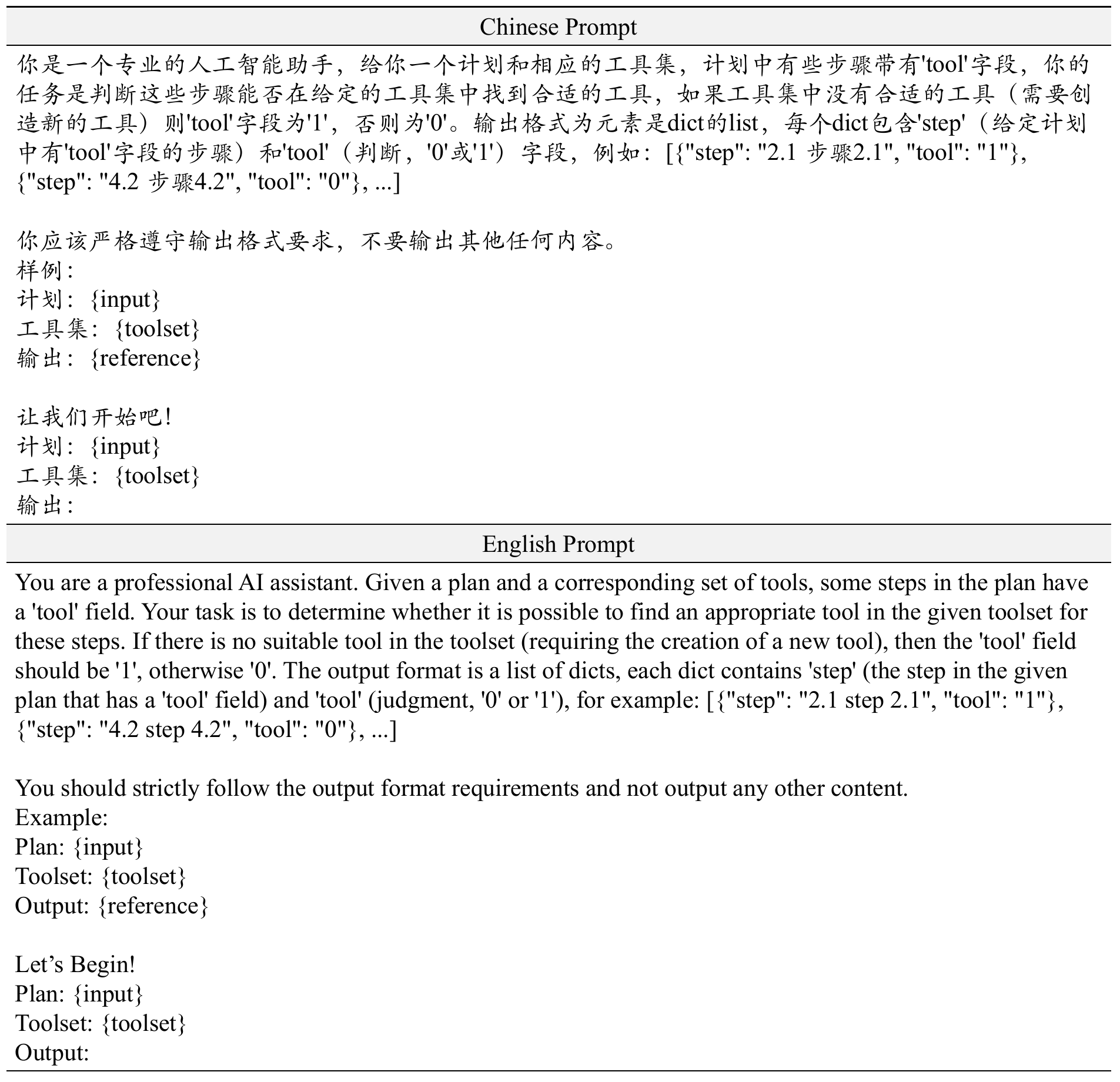}
	\caption{Prompt for tool creation awareness inference on Chinese-dataset and English-dataset.
	}
	\label{fig:tool_creation_awareness_prompt}
\end{figure*}

\begin{figure*}[t]
	\centering
	\includegraphics[width=0.95\textwidth]{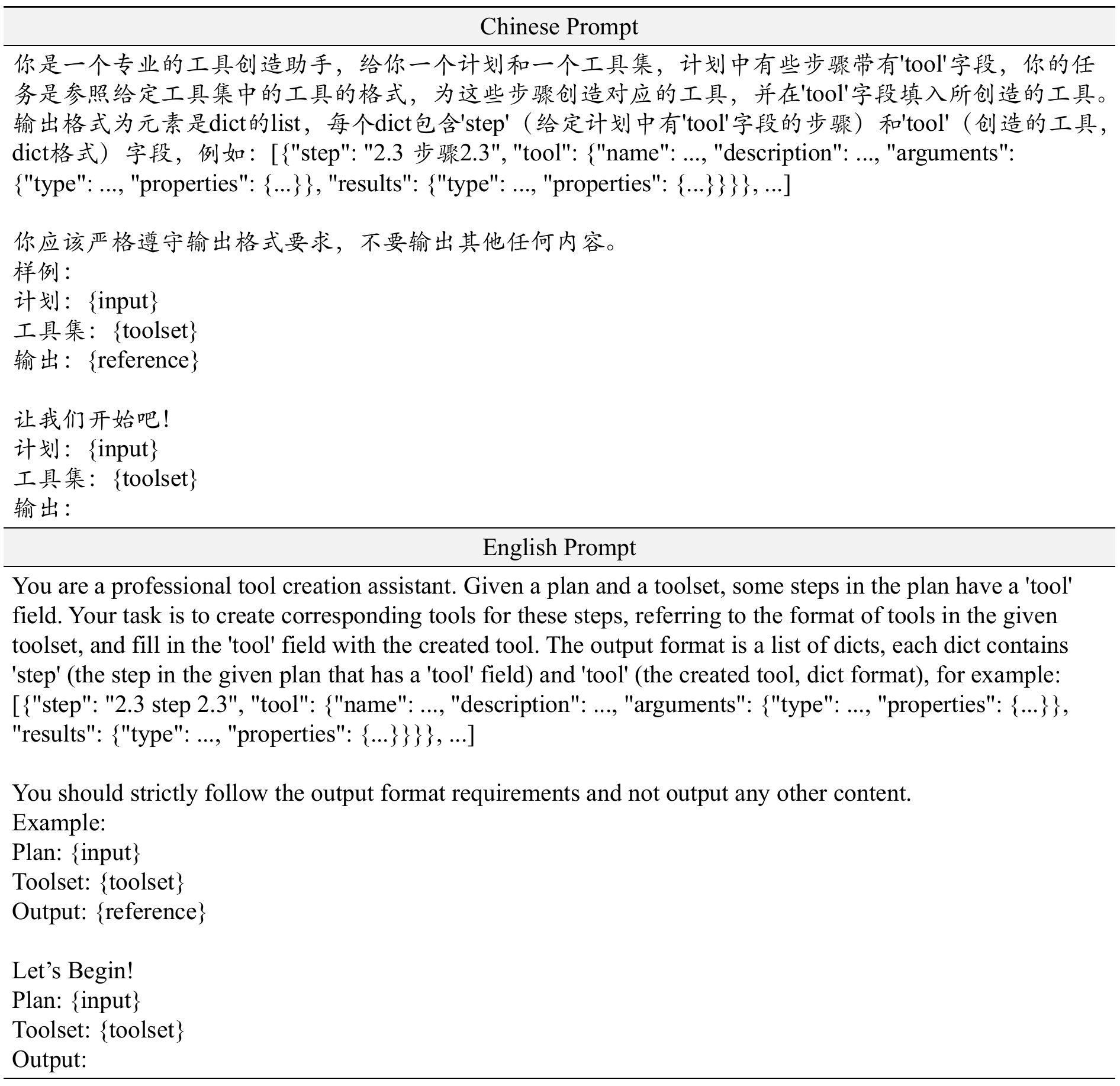}
	\caption{Prompt for tool creation inference on Chinese-dataset and English-dataset.
	}
	\label{fig:tool_creation_prompt}
\end{figure*}

\begin{figure*}[t]
	\centering
	\includegraphics[width=0.95\textwidth]{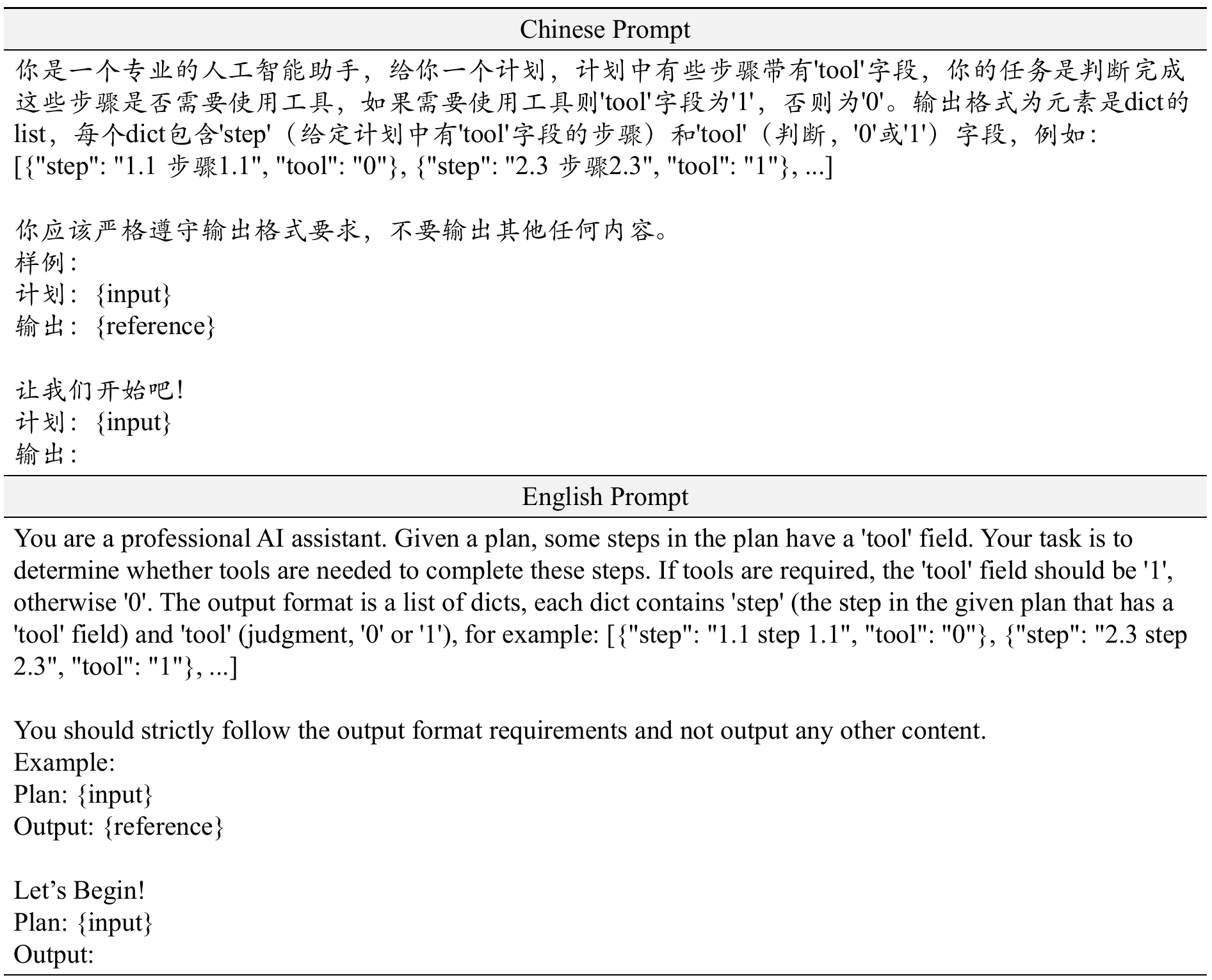}
	\caption{Prompt for tool usage awareness inference on Chinese-dataset and English-dataset.
	}
	\label{fig:tool_use_awareness_prompt}
\end{figure*}

\begin{figure*}[t]
	\centering
	\includegraphics[width=0.95\textwidth]{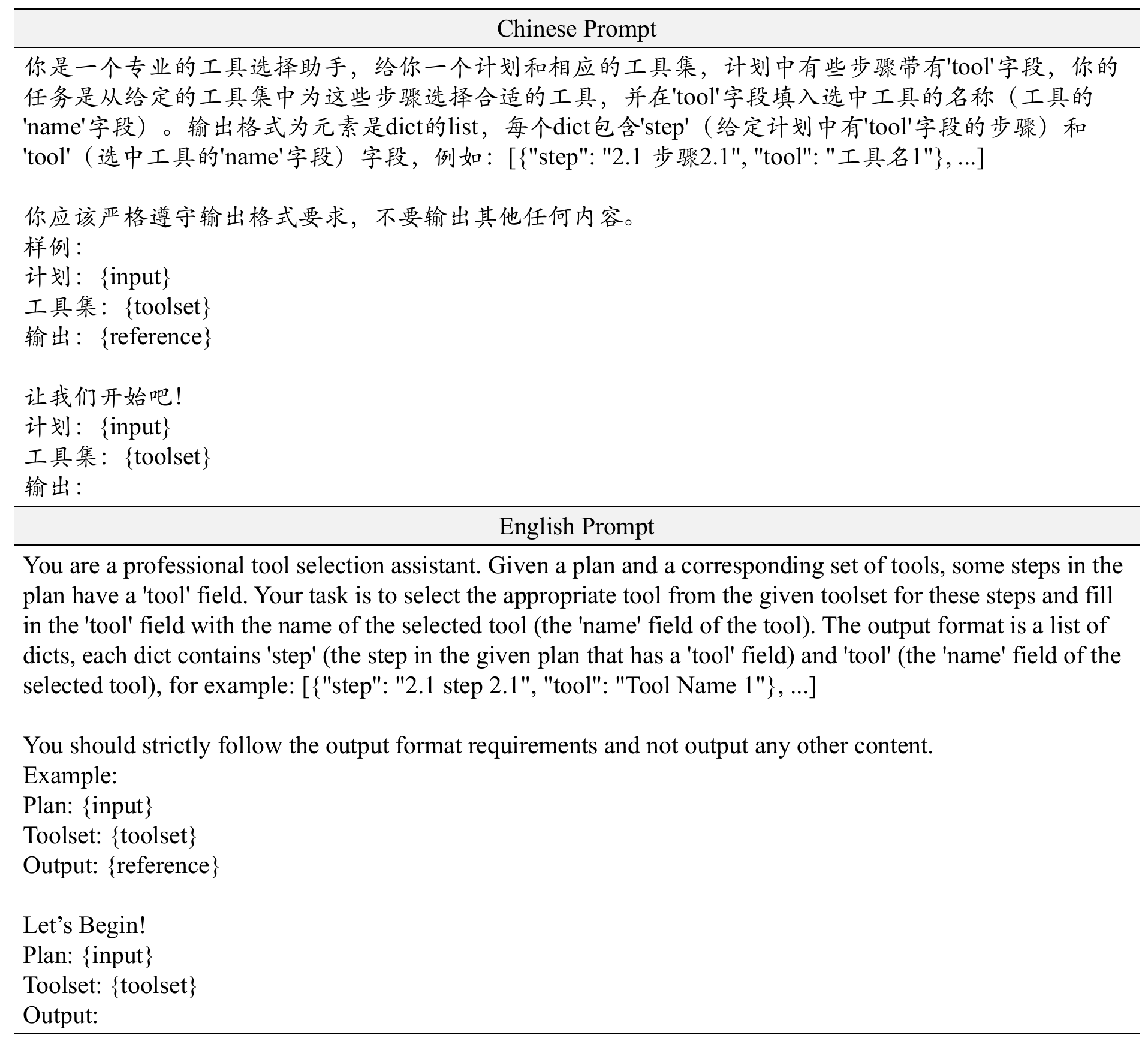}
	\caption{Prompt for tool selection inference on Chinese-dataset and English-dataset.
	}
	\label{fig:tool_selection_prompt}
\end{figure*}

\begin{figure*}[t]
	\centering
	\includegraphics[width=0.95\textwidth]{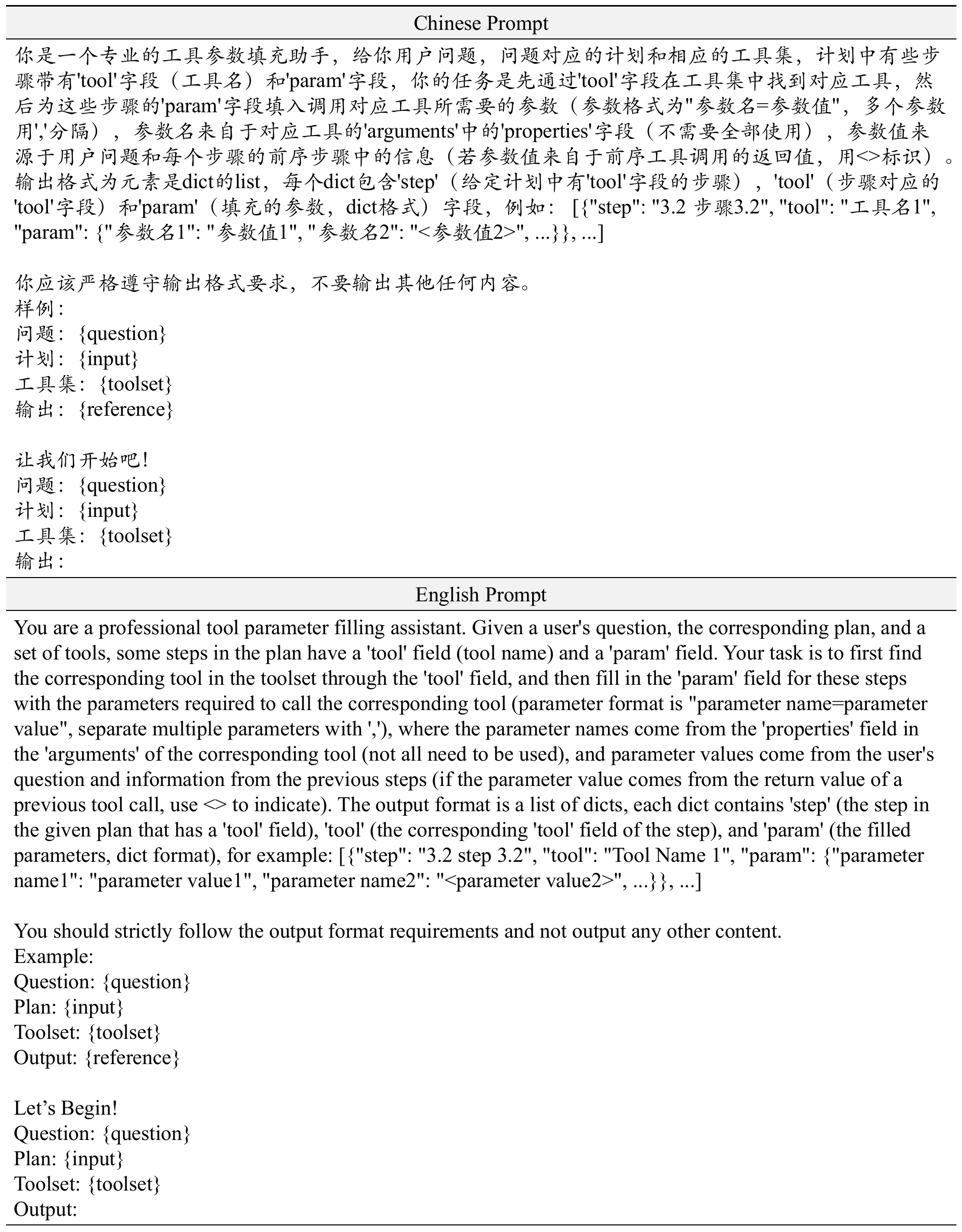}
	\caption{Prompt for tool usage inference on Chinese-dataset and English-dataset.
	}
	\label{fig:tool_usage_prompt}
\end{figure*}

During the inference of our experiments, we utilize one-shot example for all LLMs. The prompt templates can be found in \figurename~\ref{fig:planning_prompt} (planning), \figurename~\ref{fig:tool_creation_awareness_prompt} (tool creation awareness), \figurename~\ref{fig:tool_creation_prompt} (tool creation), \figurename~\ref{fig:tool_use_awareness_prompt} (tool usage awareness), \figurename~\ref{fig:tool_selection_prompt}(tool selection),
\figurename~\ref{fig:tool_usage_prompt}(tool usage), respectively.

\end{document}